\documentclass[5p,times,twocolumn]{elsarticle}

\usepackage{amsmath,amssymb,bm,mathtools}
\usepackage{graphicx}
\usepackage{booktabs,tabularx,array,multirow,makecell}
\usepackage{adjustbox}
\usepackage{enumitem}
\usepackage[protrusion=true,expansion=false]{microtype}
\usepackage{xurl}
\usepackage[colorlinks=true,allcolors=blue]{hyperref}

\graphicspath{{figures/}}
\setlist{topsep=2pt,itemsep=1pt,parsep=0pt}
\newcommand{\SI}[2]{\ensuremath{#1\,\mathrm{#2}}}
\newcommand{\second}{s}
\newcommand{\meter}{m}
\newcommand{\per}{/}
\newcommand{\squared}{^{2}}
\DeclareMathOperator*{\argmax}{arg\,max}
\DeclareMathOperator*{\argmin}{arg\,min}

\journal{Advanced Engineering Informatics}
\biboptions{sort&compress}

\makeatletter
\patchcmd{\MaketitleBox}
  {\normalsize\elsauthors\par\vskip10pt}
  {\ifx\elsauthors\@empty\else\normalsize\elsauthors\par\vskip10pt\fi}
  {}{}
\patchcmd{\MaketitleBox}
  {\footnotesize\itshape\elsaddress\par\vskip36pt}
  {\ifx\elsaddress\@empty\else\footnotesize\itshape\elsaddress\par\vskip36pt\fi}
  {}{}
\makeatother

\makeatletter
\def\ps@pprintTitle{%
  \let\@oddhead\@empty
  \let\@evenhead\@empty
  \def\@oddfoot{\hfil\thepage\hfil}%
  \let\@evenfoot\@oddfoot
}
\makeatother

\begin{document}

\begin{frontmatter}

\title{SSP: An Event-Matched Syn2Sim2Phy Cross-Domain Evaluation Framework for Autonomous-Driving VLA Models}

\author[aff1]{Haojie Feng}
\author[aff1]{Peizhi Zhang\corref{cor1}}
\ead{zhangpeizhitom@126.com}
\author[aff1]{Xinrui Zhang}
\author[aff1]{Zhuoren Li}
\author[aff1]{Junpeng Huang}
\author[aff1]{Xiurong Wang}
\author[aff2]{Dongxiao Yin}
\author[aff3]{Yuxiang Zhang}
\author[aff4]{Junfan Zhu}
\author[aff1]{Lu Xiong}

\cortext[cor1]{Corresponding author.}

\affiliation[aff1]{
    organization={College of Automotive and Energy Engineering, Tongji University},
    city={Shanghai},
    postcode={201804},
    country={China}
}

\affiliation[aff2]{
    organization={Tongji Automotive Design \& Research Institute Co., Ltd.},
    city={Shanghai},
    postcode={201804},
    country={China}
}

\affiliation[aff3]{
    organization={Hubei Jingchu Humanoid Robot Co., Ltd.},
    city={Wuhan},
    state={Hubei},
    country={China}
}

\affiliation[aff4]{
    organization={University of Chicago},
    city={Chicago},
    state={IL},
    postcode={60637},
    country={USA}
}

\begin{abstract}
Vision--language--action (VLA) models for autonomous driving jointly produce scene interpretations, language-based reasoning, and driving trajectories. Existing evaluations, however, commonly use independently selected synthetic, simulated, or physical data, thereby confounding domain shift with changes in scenario content. Consequently, observed performance gaps may reflect differences in interaction difficulty rather than genuine domain sensitivity, leading to biased cross-domain comparisons and unreliable localization of VLA capability failures. We propose SSP (Synthetic–Simulation–Physical), an event-matched Syn2Sim2Phy evaluation framework that anchors cross-domain comparison to one safety-critical interaction instance. Starting from a synthetic long-tail video, SSP establishes a validated event specification that records the interaction-defining evidence required for cross-domain comparison, including road topology, participant roles, relative motion, conflict evolution, passing order, response constraints, and event phases with a vision–language model. Platform-specific realizations are then constructed in CARLA and on a closed proving ground, and are admitted to evaluation only after transfer audits confirm that the mandatory event properties remain preserved. SSP further maps heterogeneous outputs from OpenEMMA, LLaViDA, and Alpamayo-R1 into closed semantic slots and a common \SI{1}{\second} trajectory window, enabling diagnosis of output validity, effective semantics, critical-interaction recognition, trajectory quality, and risk response. The reported Cut-in and vulnerable-road-user crossing cases yield macro-averaged Integrated VLA Capability Scores (IVCSs) of 0.259, 0.291, and 0.325 in the Synthetic, Simulation, and Physical domains, respectively, while Simulation ranks first for Cut-in, demonstrating scenario-dependent domain effects. Alpamayo-R1, OpenEMMA, and LLaViDA obtain IVCSs of 0.405, 0.338, and 0.131. Under a common OpenEMMA-style interface, Qwen3-VL-30B-A3B achieves a trajectory-quality success rate of 0.630 and an IVCS of 0.398 with approximately 3B active parameters. SSP therefore provides a reproducible scene-transfer chain and an evidence-qualified evaluation of the complete VLA behavior chain without presuming that the Physical domain is universally superior.
\end{abstract}


\end{frontmatter}

\section{Introduction}
\label{sec:introduction}

\subsection{The evaluation shift from end-to-end driving to cross-domain VLA assessment}
\label{subsec:motivation}

End-to-end autonomous driving is evolving from a direct image-to-control mapping into a vision--language--action (VLA) paradigm that combines visual understanding, language reasoning, maneuver decisions, and continuous action prediction \cite{shao2024lmdrive,sima2024drivelm,mao2023gptdriver,tian2024drivevlm,pan2024vlp,wang2025omnidrive,jiang2024senna,marcu2024lingoqa,chen2024endtoend,bojarski2016end}. OpenEMMA connects scene description, planning reasoning, and speed--curvature control through an open multimodal interface \cite{xing2025openemma}. LLaViDA jointly models explicit reasoning, discrete actions, and future trajectories \cite{liu2025llavida}, whereas Alpamayo-R1 further couples language reasoning with continuous action prediction for long-tail driving \cite{nvidia2025alpamayo}. Consequently, a VLA system no longer outputs only a steering command, a control vector, or a set of trajectory points. Its observable behavior forms a chain that includes perceiving traffic elements, understanding their interactions, assessing risk, committing to an action, and generating future motion.

This change in the evaluation target exposes limitations in conventional autonomous-driving metrics. Detection or segmentation accuracy characterizes selected perception tasks, and average or final displacement errors characterize geometric deviation from a reference trajectory. A VLA model, however, can fail independently at multiple stages. It may return a well-formed but factually incorrect description, or correctly identify a hazard and state an intention to decelerate while its trajectory continues to accelerate toward the conflict area. In safety-critical interactions such as vehicle Cut-in and vulnerable-road-user (VRU) crossing, text similarity alone cannot establish whether the participant relationship was understood, and trajectory error alone cannot determine whether the action followed a correct scene interpretation. VLA evaluation must therefore examine semantic evidence, critical-interaction understanding, explicit action commitment, trajectory feasibility, text--trajectory agreement, and sustained response after risk becomes observable.

The test data also determine the validity boundary of any evaluation conclusion. Inputs used for VLA assessment can be broadly divided into Synthetic, Simulation, and Physical domains. Synthetic data can expand long-tail events that are difficult to collect or reproduce safely; AVD2 and its EMM-AU resource demonstrate the generation of accident videos and associated knowledge evidence from accident descriptions \cite{li2025avd2}. Simulation data allow road topology, participants, environmental conditions, and triggers to be configured and replayed in platforms such as CARLA and Scenic \cite{dosovitskiy2017carla,fremont2019scenic,dreossi2019verifai}. Physical data preserve real camera imaging, vehicle motion, object-target execution error, and communication and control disturbances. A cloud-controlled closed proving ground can therefore provide high-risk interaction evidence within explicit safety boundaries \cite{zhang2025troublemaker}. The three domains primarily support long-tail coverage, controlled reproduction, and physical confirmation, respectively, and can in principle form a complementary test hierarchy.

Existing studies nevertheless tend to select samples independently from the three domains under the same nominal scenario label, without verifying whether the samples instantiate the same interaction event. Two videos labeled ``Cut-in'' may differ in road geometry, the target vehicle's initial lane, intrusion direction, relative speed, trigger time, conflict location, and risk evolution. The resulting model difference then mixes visual domain shift with scenario-content variation. Conversely, visually dissimilar videos can form a more interpretable cross-domain comparison if their road topology, participant roles, relative motion, passing order, and conflict evolution are preserved. The first requirement for cross-domain VLA evaluation is therefore not a more elaborate aggregate score, but a reliable transfer of one safety-critical event across domains together with traceable evidence of event identity.

It is likewise misleading to reduce Synthetic, Simulation, and Physical data to a monotonically increasing scale of realism. The Physical domain contains a more complete sensing and execution chain, but clutter, occlusion, reflection, target-scale variation, and execution error may make recognition and planning more difficult. Regularized lane boundaries and stable target motion in simulation can be closer to a model's training distribution. Synthetic videos can contain texture or motion artifacts but also cover extreme events that cannot be implemented safely on a proving ground. Domain value should therefore be established under shared scenario semantics and a common evaluation protocol, rather than assumed from nominal realism.

\begin{figure*}[!t]
    \centering
    \includegraphics[width=\textwidth]{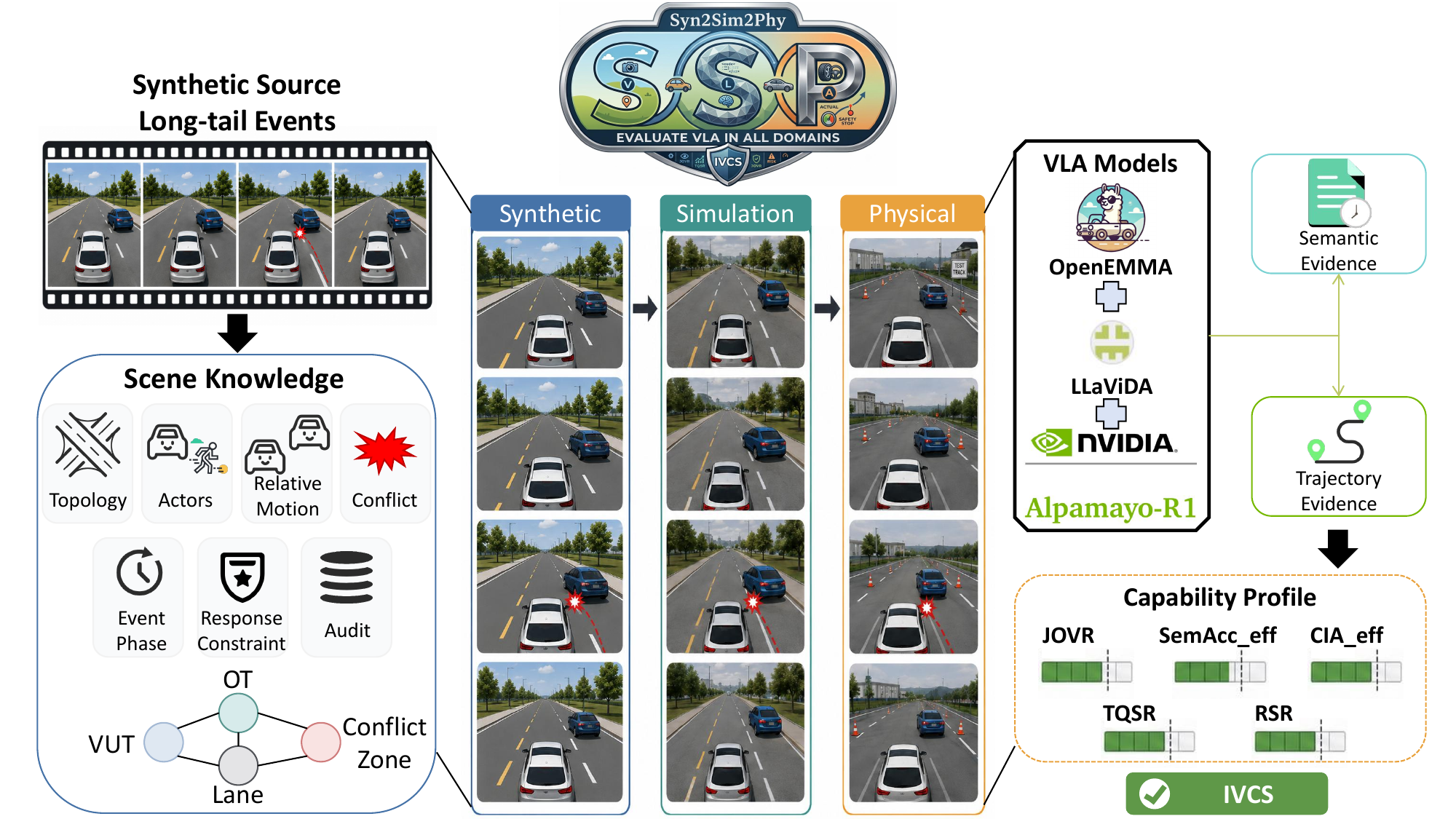}
    \caption{Conceptual framework of SSP for event-matched Syn2Sim2Phy cross-domain evaluation of autonomous-driving VLA models.}
    \label{fig:concept}
\end{figure*}

\subsection{SSP: event-matched Syn2Sim2Phy transfer and unified evaluation}
\label{subsec:ssp-overview}

To separate scenario-content variation from domain shift, we propose SSP (Synthetic--Simulation--Physical), an event-matched Syn2Sim2Phy evaluation framework for autonomous-driving VLA models (Fig.~\ref{fig:concept}). SSP does not independently search for similar cases in each domain. Instead, it uses one explicit synthetic long-tail event as the common source, performs semantic abstraction and CARLA reconstruction, and then executes the event on a closed proving ground. The resulting Synthetic, Simulation, and Physical forward-view videos are all traceable to the same road topology, participant roles, relative motion, conflict anchor, and event phases. The controlled object of comparison is thus narrowed from a broad scenario class to an auditable interaction event, making model differences more attributable to observation and execution domains.

The Synthetic-to-Simulation-to-Physical direction does not imply an ordering of data reliability. Instead, it provides a controlled construction path for an event-matched evaluation set. The Synthetic domain supplies the source evidence for the interaction to be evaluated. The Simulation domain provides a controllable realization in which topology, participant routes, conflict evolution, and event timing can be inspected with privileged state information. The Physical domain then introduces real sensing, vehicle motion, communication, tracking error, and execution disturbance. The role of these stages is therefore not to progressively generate a “more realistic” scenario, but to expose the same interaction specification to increasingly different observation and execution conditions.

SSP deliberately treats domain construction as an implementation layer rather than as a new scenario-generation algorithm. It does not require a particular semantic-to-map retrieval method, trajectory generator, or interaction controller. Instead, a domain realization is qualified by whether it preserves a predefined set of event-defining properties and satisfies the execution constraints of the target platform. In the present implementation, CARLA and the cloud-controlled proving-ground platform provide the two downstream realizations, while the transfer record preserves the correspondence among topology, participant roles, dominant relative motion, conflict region, passing order, and observable event phases \cite{tian2026video2track,zhang2025troublemaker}. This distinction is important because the methodological object of SSP is the matched cross-domain evidence set rather than the mapping algorithm used to construct any individual domain asset. Once an asset passes the transfer audit, VLA performance differences can be interpreted against changes in the observation and execution domain instead of uncontrolled changes in event content.

Three coupled evaluation problems must therefore be addressed. First, event identity must be defined independently of domain-specific appearance and execution details, so that the properties required to remain invariant can be distinguished from those allowed to vary. Second, domain-specific frame rates, durations, vehicle speeds, and event rhythms can make direct frame- or clock-time comparison interpret temporal deformation as behavioral difference; observations must instead be aligned through common event phases such as approach, trigger, conflict, and resolution. Third, VLA models differ in text templates, trajectory coordinates, sample rates, and prediction horizons, requiring their outputs to be represented through common semantic slots, actions, ego-centric coordinates, and a shared prediction interval.

Accordingly, this study addresses three research questions (RQs):
\begin{enumerate}[label=\textbf{RQ\arabic*:},leftmargin=*]
    \item Under fixed event semantics and evaluation rules, how do Synthetic, Simulation, and Physical inputs affect output validity, semantic understanding, critical-interaction recognition, trajectory quality, and risk response? Is any domain ordering stable across scenarios?
    \item Under a common behavior representation and evidence-constrained evaluation protocol, what capability-chain differences arise across heterogeneous VLA systems with different output interfaces and modeling paradigms, and at which stage does each system first fail?
    \item Under a fixed OpenEMMA-style input--output protocol, how do foundation-model scale and Dense versus mixture-of-experts (MoE) architectures affect semantic capability, trajectory capability, and active-parameter efficiency?
\end{enumerate}

\subsection{Research objective and contributions}
\label{subsec:contributions}

The objective is not to produce a leaderboard that treats the Physical domain as ground truth, nor to propose a new accident-video generator, causal-discovery method, or vehicle controller. Rather, SSP controls scenario content, traces domain conversion, and diagnoses failure along the VLA behavior chain. It changes the comparison from three independently selected videos with the same label to three implementations that alter observation and execution conditions while preserving the interaction event as far as the target platforms allow.

The principal contributions are as follows:
\begin{enumerate}[leftmargin=*,label=(\arabic*)]
    \item We introduce an event-matched Syn2Sim2Phy evaluation framework that separates event identity from domain realization. SSP defines a traceable event specification containing the topology, participant roles, dominant relative motion, conflict relationship, passing order, admissible response, and event phases that must remain invariant across domains, while allowing domain-dependent appearance, timing, dynamics, and execution disturbances to vary. Simulation and physical assets enter VLA evaluation only after explicit transfer-qualification checks, providing a controlled basis for attributing performance differences to domain conditions rather than scenario-content mismatch.
    \item We establish a common behavior representation and evidence-constrained evaluation protocol for heterogeneous VLA outputs. Free text is mapped to closed slots for objects, position, motion, conflict phase, and longitudinal and lateral actions, with each judgment grounded in explicit source text. Trajectories are converted to a common ego-centric coordinate system and a shared \SI{1}{\second} interval. Coverage, conditional accuracy, and missing-as-zero effective accuracy distinguish abstention, interface failure, and incorrect content.
    \item We construct an 18-cell matrix spanning two safety-critical scenarios, three domains, and three VLA paradigms, and further compare Dense and MoE foundation models under a common OpenEMMA-style protocol. Scenario-stratified and capability-decomposed analyses delimit the conclusions: domain value and model advantage depend on the scenario and on the stage of the behavior chain.
\end{enumerate}

\section{Related work}
\label{sec:related-work}

\subsection{From end-to-end driving to vision--language--action models}
\label{subsec:related-vla}

Early end-to-end driving learned a direct mapping from sensor observations to steering, throttle, or braking \cite{bojarski2016end}, later extending to multimodal fusion, trajectory planning, and closed-loop decision making \cite{chen2024endtoend}. Although these systems reduce handcrafted interfaces, perception, prediction, and planning are compressed into latent representations. Evaluation based on control error, trajectory deviation, route completion, or collision rate cannot localize whether a failure begins with an omitted target, an incorrect relationship, a flawed risk judgment, or an unsuitable action.

Large language and vision--language models introduce explicit semantic and reasoning interfaces. LMDrive combines language navigation, multimodal perception, and closed-loop control \cite{shao2024lmdrive}; DriveLM organizes traffic entities and relationships through graph visual question answering \cite{sima2024drivelm}; and GPT-Driver formulates trajectory planning as language modeling \cite{mao2023gptdriver}. DriveVLM and VLP connect scene understanding, driving reasoning, and trajectory planning \cite{tian2024drivevlm,pan2024vlp}, while OmniDrive and Senna extend spatiotemporal reasoning, counterfactual analysis, and end-to-end action \cite{wang2025omnidrive,jiang2024senna}. LingoQA further shows that driving-language evaluation must focus on traffic facts and decision evidence instead of lexical overlap \cite{marcu2024lingoqa}.

The three systems evaluated here represent distinct capability pathways. OpenEMMA follows the multimodal design of EMMA and uses an open foundation model plus structured prompting to produce scene interpretation, driving intent, and speed--curvature control \cite{xing2025openemma,hwang2024emma}. LLaViDA jointly generates an explanation, an action, and a future trajectory after driving-task supervision and trajectory preference optimization \cite{liu2025llavida}. Alpamayo-R1 connects reasoning and continuous action prediction for long-tail cases \cite{nvidia2025alpamayo}. Because their histories, text templates, coordinate systems, sample rates, and horizons differ, raw outputs are not directly comparable. Evaluation requires a shared semantic, action, and trajectory representation that still preserves native model capability and exposes missing or invalid interfaces.

\subsection{Functional roles of synthetic, simulated, and physical data}
\label{subsec:related-domains}

Real-world datasets provide the main foundation for autonomous-driving perception, prediction, interaction understanding, and planning. nuScenes, the Waymo Open Motion Dataset, and Argoverse 2 provide large-scale multimodal perception and motion-forecasting resources  \cite{caesar2020nuscenes,ettinger2021waymo,wilson2021argoverse2}, while the Interactive Enhanced Driving Dataset (IEDD) further concentrates interaction-rich trajectory data and aligns them with structured semantics and language supervision for VLA-oriented learning and evaluation \cite{feng2026iedd}. nuPlan connects recorded driving with closed-loop planning evaluation \cite{caesar2021nuplan}; NAVSIM and Bench2Drive advance log-based planning and multi-capability closed-loop assessment \cite{dauner2024navsim,jia2024bench2drive}. These resources are large and diverse, but do not generally instantiate a selected safety-critical event simultaneously as synthetic video, controlled simulation, and physical proving-ground video.

Synthetic data can expand rare events at low cost. AVD2 generates accident videos and semantic evidence from accident descriptions \cite{li2025avd2}, while domain-randomization and synthetic-data studies show how changes in texture, illumination, object appearance, and environment can enlarge coverage \cite{tobin2017domain,tremblay2018training,richter2016playing,ros2016synthia}. Generated video, however, may exhibit object deformation, identity drift, inconsistent road relationships, or temporal discontinuity. Such defects can change the event itself for a VLA model, so synthetic samples require explicit audits of roles, relative motion, conflict relationships, and phases.

Simulation is better suited to controlled repetition and diagnosis. CARLA exposes maps, vehicles, pedestrians, environments, and sensors \cite{dosovitskiy2017carla}; Scenic supports probabilistic scenario specification \cite{fremont2019scenic}; and VerifAI and CommonRoad support formal analysis and composable motion-planning cases \cite{dreossi2019verifai,althoff2017commonroad}. Simulation can fix topology, paths, and triggers while retaining privileged state logs, but its rendering, behavior models, and dynamics approximate reality. It can establish behavior under controlled conditions, but cannot alone establish that behavior under physical sensing and execution.

Physical data preserve real imaging, vehicle motion, tracking error, and environmental disturbance. Closed proving grounds can stage Cut-in and VRU events with monitoring and emergency stops \cite{zhang2025troublemaker,koopman2016challenges}, but topology, device capability, safety limits, and reset cost constrain coverage. Physical tests are therefore most useful as confirmation after screening rather than as a large-scale source of long-tail diversity. The three domains are complementary: Synthetic expands coverage, Simulation provides controlled reproduction, and Physical supplies execution-level confirmation.

\subsection{Cross-domain event consistency and behavior-chain evaluation}
\label{subsec:related-transfer}

Automated-driving scenarios are commonly described at functional, logical, and concrete levels \cite{ulbrich2015scene,menzel2018scenarios,bagschik2018ontology,feng2020library}. A video contains road, participant, and interaction evidence, but does not directly provide map identifiers, spawn points, paths, trigger windows, or safety constraints. Converting video into an executable task should therefore extract stable topology, roles, relative motion, conflict relationships, and event phases before parameterizing the target platform. Asking a monocular VLM to invent absolute speeds, distances, and a complete executable script can yield a formally valid task whose semantics no longer match the source.

For cross-domain evaluation, however, constructing an executable realization is only part of the problem. A more fundamental question is whether the resulting assets still constitute the same interaction event. Pixel similarity is neither necessary nor sufficient: changes in texture, rendering, speed, or local geometry may be acceptable, whereas changes in participant role, dominant relative motion, conflict relationship, passing order, or event evolution can invalidate a domain comparison. Cross-domain evaluation therefore requires an explicit separation between event-defining invariants and domain-dependent variables, together with an auditable criterion for accepting or rejecting each realization.

Structured scene representations provide a practical means of recording such invariants. Tuples, scene graphs, and temporal interaction graphs can represent relationships among roads, participants, conflict regions, and event phases \cite{pearl2009causality,scholkopf2021causal,battaglia2018relational}. In SSP, the representation is used as an evidence and qualification record rather than as a scenario-generation model: it records what must remain unchanged, what may legitimately vary, and which source evidence supports each field. Platform-specific configuration is then treated as an implementation step whose output must satisfy this common event specification.

Physical execution additionally introduces state-dependent triggering, object-target tracking, communication disturbance, and safety supervision \cite{zhao2018accelerated,feng2020library,feng2023dense}. The cloud-controlled functions provided by Real-World Troublemaker \cite{zhang2025troublemaker} are used here to realize and record the Physical-domain asset. Their role in SSP is not to generate adversarial scenarios, but to provide an observable physical realization whose correspondence to the Synthetic and Simulation assets can be audited before VLA evaluation.

DeepTest and DeepRoad test autonomous-driving robustness through image transformation and generative domain transfer \cite{tian2018deeptest,zhang2018deeproad}; SafeBench evaluates closed-loop behavior in safety-critical scenario libraries \cite{xu2022safebench}. These approaches emphasize final control, task completion, or incident outcomes and do not localize failures in a VLA system that jointly emits language and trajectories. Open-ended LLM judges such as G-Eval can support text evaluation \cite{liu2023geval}, but unconstrained grading is sensitive to prompt form, verbosity, and style and does not establish language--trajectory consistency. SSP instead restricts an LLM to evidence extraction from a closed label set and delegates correctness to deterministic scoring after trajectory normalization.

In summary, prior work does not jointly provide (i) a continuous Syn2Sim2Phy chain from one long-tail event, (ii) auditable qualification of event identity across domain realizations, (iii) a common behavior representation for heterogeneous VLA language and trajectory outputs, and (iv) a matched design that studies domain, VLA paradigm, and foundation-model architecture. SSP addresses this combined gap by linking semantic extraction, simulator reconstruction, proving-ground execution, and behavior-chain evaluation in one traceable protocol.

\section{Method: Event-identity-preserving Syn2Sim2Phy evaluation}
\label{sec:method}

\subsection{Problem definition and overall framework}
\label{subsec:problem}

Let \(\mathcal S\) denote a set of synthetic safety-critical scenarios, \(\mathcal D=\{\mathrm{Syn},\mathrm{Sim},\mathrm{Phy}\}\) the three data domains, and \(\mathcal M\) the VLA models under evaluation. SSP does not attempt to generate three pixel-identical videos. Its objective is to preserve the interaction-defining properties of each event while changing the observation and execution domain, and then to evaluate how the VLA behavior chain responds. In this study, a scenario denotes a functional interaction class, whereas an event instance denotes a concrete realization of that class. Cross-domain matching is therefore imposed at the event-instance level. Event matching does not require identical texture, background, absolute speed, or pointwise trajectory; it requires the preservation of a common, inspectable, and traceable set of interaction-defining properties, including road topology, participant roles, dominant relative motion, conflict type, passing order, and event phases.

For scenario \(s\in\mathcal S\), SSP begins with a synthetic source video \(V_s^{\mathrm{syn}}\). Hierarchical semantic extraction and human review produce validated scene knowledge \(K_s^{\star}\), which is compiled into CARLA and closed-track artifacts. The transfer is
\begin{equation}
V_s^{\mathrm{syn}}
\xrightarrow{f_{\mathrm{sem}}} K_s^{\star}
\xrightarrow{f_{\mathrm{sim}}}
\left(\mathcal A_s^{\mathrm{sim}},V_s^{\mathrm{sim}}\right)
\xrightarrow{f_{\mathrm{phy}}}
\left(\mathcal A_s^{\mathrm{phy}},V_s^{\mathrm{phy}}\right),
\label{eq:transfer}
\end{equation}
where \(f_{\mathrm{sem}}\) combines semantic extraction, rule checks, and human verification; \(\mathcal A_s^{\mathrm{sim}}\) and \(\mathcal A_s^{\mathrm{phy}}\) contain executable configurations, control parameters, and state logs; and \(V_s^{\mathrm{sim}}\) and \(V_s^{\mathrm{phy}}\) are the forward-view videos supplied to the VLA model. Equation~\eqref{eq:transfer} represents continuous compilation rather than independent reconstruction. The Simulation compiler consumes the validated \(K_s^{\star}\), and the Physical compiler consumes the same version plus the anchor relationships validated in simulation.

After event-phase alignment, each domain video is processed by model \(m\in\mathcal M\):
\begin{equation}
Y_{s,d,m}=F_m\!\left(V_s^d;Q_m,H_m\right)
=\left(Y_{s,d,m}^{\mathrm{text}},Y_{s,d,m}^{\mathrm{traj}}\right),
\qquad d\in\mathcal D,
\label{eq:model-output}
\end{equation}
where \(Q_m\) is the minimally adapted task prompt and \(H_m\) records the video and control histories, coordinates, output format, sample interval, and decoding configuration. Fixing \(K_s^{\star}\), the input clip, and the evaluator separates scenario, domain, and model factors in the experimental organization; it does not assume that the domain effect is constant across scenarios.

\begin{figure*}[!t]
    \centering
    \includegraphics[width=\textwidth]{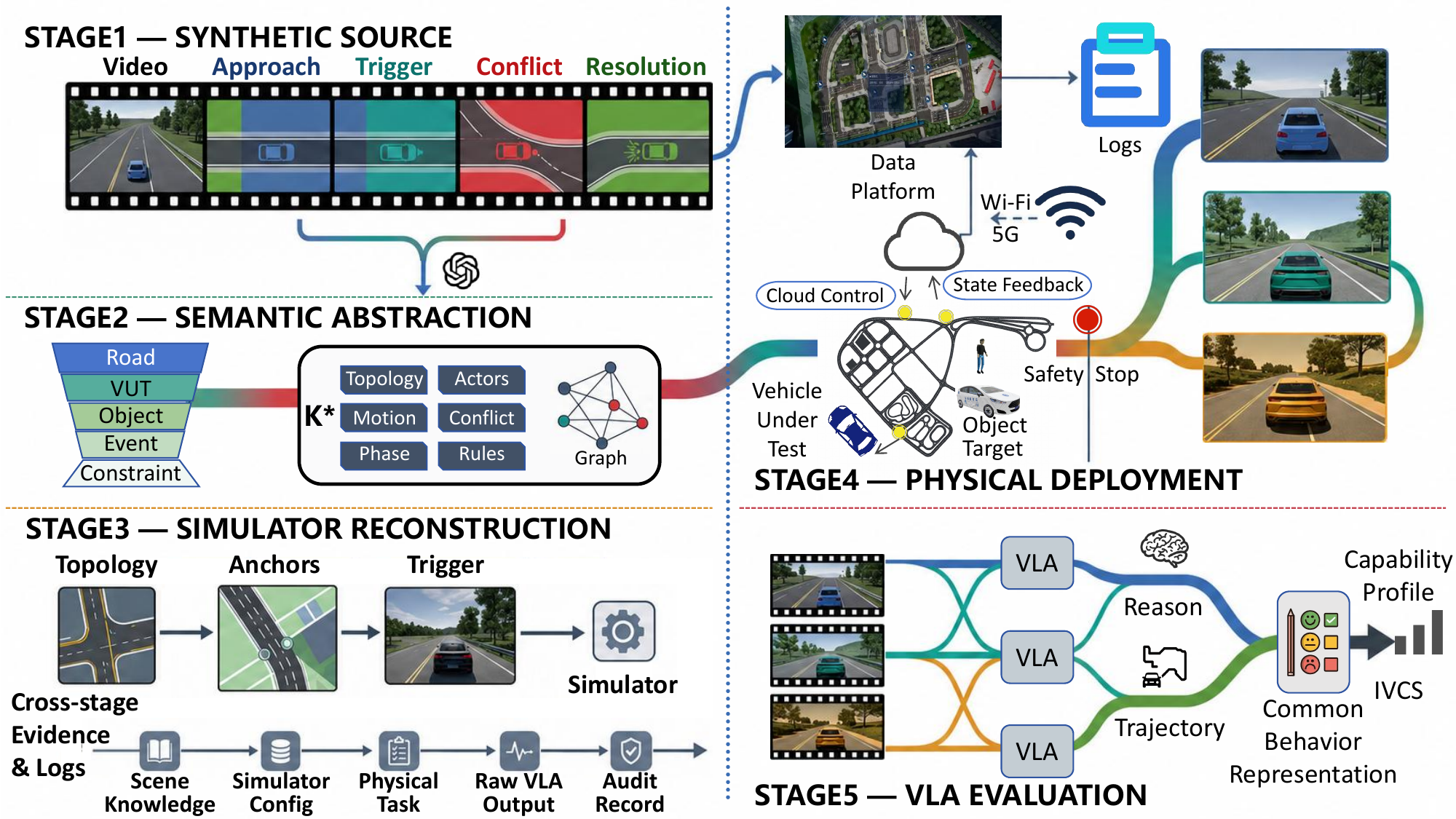}
    \caption{Overall SSP workflow from synthetic source scenes to physical deployment and unified VLA evaluation.}
    \label{fig:workflow}
\end{figure*}

Each scenario forms a versioned evidence package
\begin{equation}
\mathcal P_s=\left\{V_s^{\mathrm{syn}},K_s^{\star},\mathcal A_s^{\mathrm{sim}},V_s^{\mathrm{sim}},
\mathcal A_s^{\mathrm{phy}},V_s^{\mathrm{phy}},\mathcal E_s,L_s\right\},
\label{eq:evidence-package}
\end{equation}
where \(\mathcal E_s\) stores shared event-phase labels and evidence times, and \(L_s\) records field provenance, human edits, compiler versions, random seeds, control parameters, execution deviations, and on-site changes. The methodological output is therefore not a new vehicle controller or an isolated scene-graph algorithm, but a linked set of three-domain assets, platform artifacts, and audit records. A semantic, trajectory, or risk-response score can be traced backward through the original model output, domain video, execution log, and validated scene knowledge. The complete compilation, audit, and evaluation workflow is summarized in Fig.~\ref{fig:workflow}.

\subsection{Source-event specification and identity constraints}
\label{subsec:semantic-abstraction}

SSP uses a synthetic safety-critical video as the common scenario source. We select assets from the AVD2 accident-video generation framework and its EMM-AU resource that have a clear interaction structure, can be reproduced in CARLA and on the proving ground, and satisfy physical safety constraints \cite{li2025avd2}. The current experiments cover vehicle Cut-in and VRU crossing. Cut-in is defined by a vehicle intruding laterally into the vehicle under test (VUT) lane; VRU crossing is defined by a pedestrian or cyclist entering the VUT's travel corridor. Selection is based on identifiable roles, road relationships, conflict regions, onset and resolution of risk, and defensible longitudinal or lateral responses, rather than on visual severity.

Each source video is cropped to include pre-risk, approach, conflict exposure, and response evolution, and is labeled manually at nine time points. The labels record the target, relative position, motion trend, conflict phase, and allowable longitudinal and lateral responses. Generation source, original version, and crop metadata are retained. A clip is rejected if it exhibits material object deformation, participant identity drift, discontinuous critical motion, inconsistent road relationships, or an unidentifiable conflict event. This quality gate prevents a semantic defect in generated video from propagating into simulation and physical execution.

The source video is converted into an event identity specification whose purpose is to support cross-domain control rather than to generate an executable scenario directly. SSP therefore uses a global-to-local, semantic-to-event prompt hierarchy. It first identifies road type, lane count and direction, boundaries, and feasible maneuvers; then the VUT region, heading, and intent; then the class, role, initial relative position, and motion change of each object target (OT); and finally the interaction type, passing order, conflict region, normative response, and the approach, trigger, conflict, and resolution phases. For source video \(V_s^{\mathrm{syn}}=\{I_t\}_{t=1}^{T}\), the hierarchical VLM produces
\begin{equation}
\widetilde K_s=F_{\mathrm{VLM}}\!\left(V_s^{\mathrm{syn}},P_{\mathrm{hier}}\right)
=\{T_s,A_s,M_s,I_s,E_s,C_s,U_s\},
\label{eq:candidate-knowledge}
\end{equation}
where \(T_s\) denotes road and lane topology; \(A_s\) participant classes, identities, roles, and initial regions; \(M_s\) relative positions, directions, speed trends, and longitudinal/lateral changes; \(I_s\) interaction type, conflict relationship, passing order, and admissible responses; \(E_s\) event phases and evidence frames; \(C_s\) executable speed and distance ranges, camera requirements, site limits, and safety constraints; and \(U_s\) confidence, provenance, unresolved fields, and review status. Absolute speed, physical distance, or precise post-encroachment time that cannot be recovered reliably from monocular video is retained as an interval, order relation, or unresolved field rather than fabricated by the VLM.

Rule checks and human verification yield \(K_s^{\star}=\mathcal R(\widetilde K_s)\). Review is restricted to safety-critical fields, evidence sufficiency, and platform executability; the event is not reinterpreted independently in each domain. Every edit is written to \(L_s\) and assigned a unique version so that both compilers consume the same definition. For serialization and platform exchange, \(K_s^{\star}\) is represented as a typed spatial--temporal interaction graph
\begin{equation}
\mathcal G_s=\left(\mathcal V_s,\mathcal E_s^{\mathrm{sp}},\mathcal E_s^{\mathrm{tmp}},
\mathcal E_s^{\mathrm{int}},P_s,U_s\right),
\label{eq:interaction-graph}
\end{equation}
where nodes represent roads, lanes, the VUT, OTs, the conflict zone, and event phases; \(\mathcal E_s^{\mathrm{sp}}\) describes occupancy, adjacency, relative position, and path intersection; \(\mathcal E_s^{\mathrm{tmp}}\) describes phase order; \(\mathcal E_s^{\mathrm{int}}\) describes trigger, motion change, conflict exposure, and normative response; \(P_s\) stores determined values with units; and \(U_s\) stores intervals, uncertainty, and evidence provenance. The graph is an audit and packaging mechanism, not a claim of a new causal-discovery model.

The event specification is accepted for cross-domain realization only after five groups of identity and executability constraints are satisfied:
\begin{enumerate}[leftmargin=*,label=\textbf{C\arabic*:}]
    \item \textbf{Type consistency.} Participants, roads, lanes, and the conflict zone must enter valid relations, and VUT/OT identities must remain stable across phases.
    \item \textbf{Topological consistency.} The origin and target lanes of a Cut-in must be adjacent; a VRU path must intersect the VUT corridor; and participant routes must be connected on the target platform.
    \item \textbf{Temporal consistency.} The event must follow approach \(\rightarrow\) trigger \(\rightarrow\) conflict \(\rightarrow\) resolution, and conflict cannot precede an observable trigger.
    \item \textbf{Interaction consistency.} Conflict zone, passing order, and allowable responses must be compatible, and the admissible-response set cannot be empty after risk becomes observable.
    \item \textbf{Engineering executability.} Coordinates, speeds, distances, field of view, trigger windows, and safety limits must have explicit units and lie within CARLA or proving-ground limits.
\end{enumerate}
Any hard-constraint failure returns the field, evidence, and reason for revision instead of producing an apparently plausible task. The failure can thus be localized to inadequate source video, semantic extraction, a structural contradiction, or an unsatisfied platform constraint.

\subsection{Syn2Sim2Phy compilation and cross-domain consistency audit}
\label{subsec:compilation}

The Syn-to-Sim stage first matches topology and then places anchors. CARLA/OpenDRIVE maps are parsed into a topology library \(\mathcal B^{\mathrm{sim}}\) containing segment type, lane direction and connectivity, permitted maneuvers, entrances and exits, and candidate conflict areas. Retrieval considers road structure, participant paths and conflicts, and platform feasibility rather than visual similarity. The selected topology block is
\begin{equation}
b_s^{\mathrm{sim}\star}=\argmax_{b\in\mathcal B^{\mathrm{sim}}}
\left[\lambda_T S_T(K_s^{\star},b)+\lambda_I S_I(K_s^{\star},b)
+\lambda_C S_C(K_s^{\star},b)\right],
\label{eq:topology-retrieval}
\end{equation}
where \(S_T\), \(S_I\), and \(S_C\) score topology, interaction, and constraint satisfaction. We set \(\lambda_T=\lambda_I=\lambda_C=1/3\) before model evaluation and do not tune them against downstream IVCS results. The retrieved block must also pass lane-connectivity and conflict-geometry checks. When candidates are tied, a reviewer selects the block using source-video evidence; the VLM does not generate map coordinates.

For the VUT and every OT, the compiler constructs an executable start--goal--conflict anchor set
\begin{equation}
\begin{aligned}
\mathcal Q_i^d=\{&\bm p_{i,\mathrm{start}}^d,\bm p_{i,\mathrm{goal}}^d,
\bm p_{i,\mathrm{conflict}}^d,\\
&\Omega_{i,\mathrm{trigger}}^d,\Omega_{i,\mathrm{safe}}^d\},
\quad d\in\{\mathrm{Sim},\mathrm{Phy}\}.
\end{aligned}
\label{eq:anchor-set}
\end{equation}
Start and goal determine the reference route, the conflict anchor refers to the shared conflict region, \(\Omega_{i,\mathrm{trigger}}^d\) specifies a relative-distance, arrival-time, or state trigger, and \(\Omega_{i,\mathrm{safe}}^d\) contains speed, acceleration, minimum-separation, and termination limits. These anchors translate event-defining structure rather than copy an unobservable pointwise trajectory. Local speed profiles and trajectories may be calibrated within \(C_s\), but cannot change participant identity, dominant motion direction, conflict region, or passing order.

The CARLA compiler converts \(b_s^{\mathrm{sim}\star}\) and \(\mathcal Q_i^{\mathrm{Sim}}\) into a map version, participant blueprints, spawn points, reference routes, initial states, behavior triggers, and the VUT forward-camera configuration \cite{dosovitskiy2017carla}. Runs use synchronous stepping and log the random seed, weather, camera field of view, resolution, frame rate, and trigger parameters. Calibration preserves event phases, conflict location, and passing order rather than optimizing pixel similarity. The forward-view video is the only Simulation-domain input to the VLA; bird's-eye views, exact poses, and conflict geometry are reserved for labeling, task acceptance, and transfer audits.

The Sim-to-Phy stage continues from the same \(K_s^{\star}\), the proving-ground topology library, and simulation-validated anchor relationships. The physical library records available road blocks, lane width, entrances and exits, OT-reachable regions, conflict areas, safe-stop zones, and speed limits. A block-level transformation maps normalized longitudinal--lateral anchors to the proving ground:
\begin{equation}
\bm p_i^{\mathrm{phy}}=T_b\!\left(\bm p_i^{\mathrm{local}}\right).
\label{eq:physical-mapping}
\end{equation}
For a locally rigidly alignable block, \(T_b(\bm p)=R_b\bm p+\bm t_b\); for a curved segment, the mapping uses arc length and a Frenet frame along the site reference line. The transformation defines the VUT route, OT paths, conflict point, and trigger region. Local trajectories may be adjusted for site curvature, OT dynamics, tracking accuracy, and safety, but all changes are versioned and cannot alter roles, relative-motion direction, passing order, or event phases.

Physical execution uses the cloud-controlled infrastructure of Real-World Troublemaker \cite{zhang2025troublemaker}. VUT and OT states are streamed to the cloud; scenario management evaluates state triggers and dispatches OT references; local controllers track longitudinal and lateral motion and return status; and the monitor records forward video, states, trigger events, and controller acknowledgments. Communication faults, boundary violations, excessive tracking error, or a task termination condition invoke a safe stop. Although the underlying platform supports dynamic games and risk regulation, SSP uses only the triggering, trajectory dispatch, feedback, tracking, and safety functions required to construct the Physical-domain data. High-precision localization and cloud states are not VLA inputs.

The central SSP operation begins after a candidate domain realization has been constructed: the asset must be qualified against the source event before it can enter the VLA evaluation matrix. Because domain frame rates, durations, speeds, and rhythms differ, SSP aligns observable phases rather than raw frame numbers. For event \(e\in\mathcal E_s\),
\begin{equation}
\phi_{s,d,e}=\frac{t_{s,d,e}-t_{s,d,\mathrm{start}}}
{t_{s,d,\mathrm{end}}-t_{s,d,\mathrm{start}}}\in[0,1].
\label{eq:normalized-phase}
\end{equation}
The event-phase deviation (EPD) from the Synthetic source is
\begin{equation}
\mathrm{EPD}_{s,d}=\frac{1}{|\mathcal E_s|}
\sum_{e\in\mathcal E_s}\left|\phi_{s,d,e}-\phi_{s,\mathrm{Syn},e}\right|,
\label{eq:epd}
\end{equation}
and the key-property retention (KPR) for discrete topology, role, motion, interaction, and passing-order fields is
\begin{equation}
\mathrm{KPR}_{s,d}=
\frac{\sum_k w_k\,\mathbf 1\!\left(q_{s,d,k}=q_{s,\mathrm{Syn},k}\right)}
{\sum_k w_k}.
\label{eq:kpr}
\end{equation}
KPR and EPD qualify a domain asset for downstream VLA comparison; they are not VLA capability scores. All mandatory properties---topology, participant role, dominant motion, conflict type, and passing order---must pass. EPD describes temporal deformation. These quantities are protocol-level qualification measures for the paired domain assets rather than population-level estimators. Accordingly, SSP does not score how well a mapping algorithm reconstructs the source trajectory. It tests whether each resulting domain asset remains a valid realization of the same interaction event. KPR evaluates preservation of discrete event identity, whereas EPD quantifies temporal deformation among otherwise qualified realizations.

An asset enters model evaluation only if all three domains preserve the mandatory event properties, each key phase is locatable, the forward camera covers the participants and conflict region, and the log contains no safety abort, identity switch, or unrecorded on-site change. Texture, background, local speed, tracking error, and communication disturbance may vary because they are domain properties, but they must be disclosed in \(L_s\). SSP therefore specifies what must remain invariant, what may vary, and what invalidates the comparison; it does not claim equivalence over all latent variables.

\subsection{Common VLA representation, evidence extraction, and reproducibility}
\label{subsec:common-representation}

SSP does not modify model internals; it adapts only the input clip, task prompt, and output parser. Native VLA systems differ in text templates, trajectory coordinates, sampling intervals, prediction horizons, and failure modes. Each model-specific adapter therefore converts the native output into a common behavior record containing the raw text, interface-validity flags, evidence-grounded semantic slots, explicit longitudinal and lateral actions, and an ego-centric trajectory. The adapter performs parsing and coordinate conversion only. Missing outputs, shape mismatches, non-finite values, invalid timestamps, or trajectories that do not cover the required common interval are marked invalid rather than repaired through extrapolation or default values.

Free text is mapped to closed slots for object, relative position, motion pattern, conflict phase, longitudinal action, and lateral action. A trajectory is transformed to the local VUT frame \(\bm p(t)=[x(t),y(t)]^\top\) and linearly interpolated only within its valid native support:
\begin{equation}
\bm\tau=\{0,0.5,1.0\}\ \mathrm{s},\qquad
\widehat{\bm p}_{s,d,m}(\bm\tau)=
\operatorname{Interp}\!\left(\bm p_{s,d,m}(t),\bm\tau\right).
\label{eq:common-trajectory}
\end{equation}
If the native prediction does not cover \([0,1.0]\)~s, it remains invalid. Comparing only the shared support avoids giving an automatic advantage to a longer horizon or denser sampling. Nine labels in each video are matched to model outputs through monotone one-to-one dynamic programming: the lexicographic objective first maximizes the number of matches and then minimizes total time error, with a default tolerance of \SI{0.11}{\second}. At the earliest instant, only object and relative-position evidence is scored unless dynamic evidence is explicitly available.

Text evaluation follows a three-stage procedure: high-precision rules, constrained-LLM completion, and deterministic scoring. The rule parser first extracts explicit objects, positions, motions, and actions. Only when a slot is missing may the LLM choose from a predefined closed label set, and it must return a supporting span from the original model output. Negation scope, subject binding, token boundaries, synonym mapping, and label legality are then checked. The extractor cannot see the ground-truth label; deterministic code computes correctness after extraction. The LLM is thus an evidence extractor rather than an open-ended grader of fluency or verbosity.

Trajectory outputs are transformed into the local VUT coordinate frame and restricted to their valid shared prediction support. Phase-aligned event labels are then associated with model outputs through a common temporal-matching procedure. The resulting records support behavior-chain evaluation from interface validity and semantic evidence to interaction recognition, action commitment, trajectory quality, and risk response; the corresponding metrics are formally defined in Section~\ref{sec:metrics}.

All artifacts share a scenario identifier: source video, structured scene knowledge, CARLA configuration, physical task, event labels, raw VLA output, parsed evidence, score record, and plotting script. Transformations record software and model versions, seeds or control parameters, review edits, and site adjustments. Prompts, temperature, video preprocessing, and parsers are frozen with the experimental version. This versioned chain enables domain, model, and adapter effects to be inspected separately and supports later extension to additional scenarios, models, and proving grounds.

\section{Experiments and results}
\label{sec:experiments}

\subsection{Experimental design}
\label{subsec:experimental-design}

The experiment forms a \(2\times3\times3\) crossed matrix comprising two safety-critical scenarios (Cut-in and VRU crossing), three domains (Synthetic, Simulation, and Physical), and three VLA paradigms (OpenEMMA, LLaViDA, and Alpamayo-R1). RQ1 compares domains using cross-model macro averages and scenario-stratified results. RQ2 compares the selected heterogeneous VLA systems using cross-domain macro averages and scenario-stratified results. RQ3 fixes the OpenEMMA-style data flow, task definition, and parser to compare foundation-model scale and Dense/MoE architecture.

All systems receive the same task semantics, label space, and evaluation objective. Prompts are minimally adapted to native interfaces without changing the requested semantic fields, allowable actions, or common trajectory definition. Phase-aligned labels and the same deterministic scoring rules are applied to every scenario--domain--model cell. The resulting macro averages and scenario strata provide paired, protocol-level comparisons of the evaluated assets.

\begin{figure*}[!t]
    \centering
    \includegraphics[width=\textwidth]{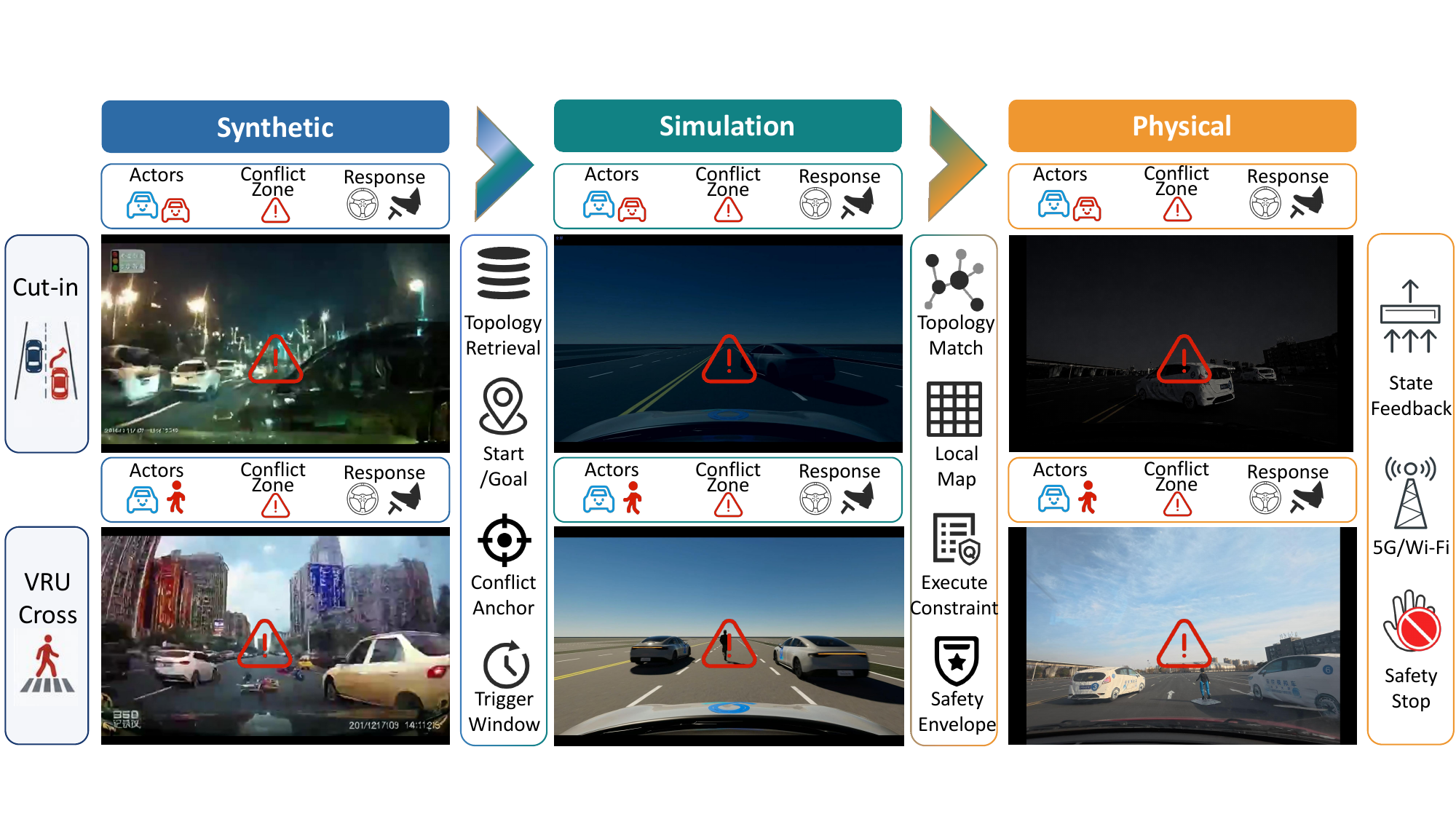}
    \caption{Event-matched Syn2Sim2Phy transfer examples for Cut-in and VRU-crossing events. Colored conflict-region annotations and other overlays are used only for visualization and are not provided as inputs to the VLA models.}
    \label{fig:transfer-examples}
\end{figure*}

\begin{figure*}[!t]
    \centering
    \includegraphics[width=\textwidth]{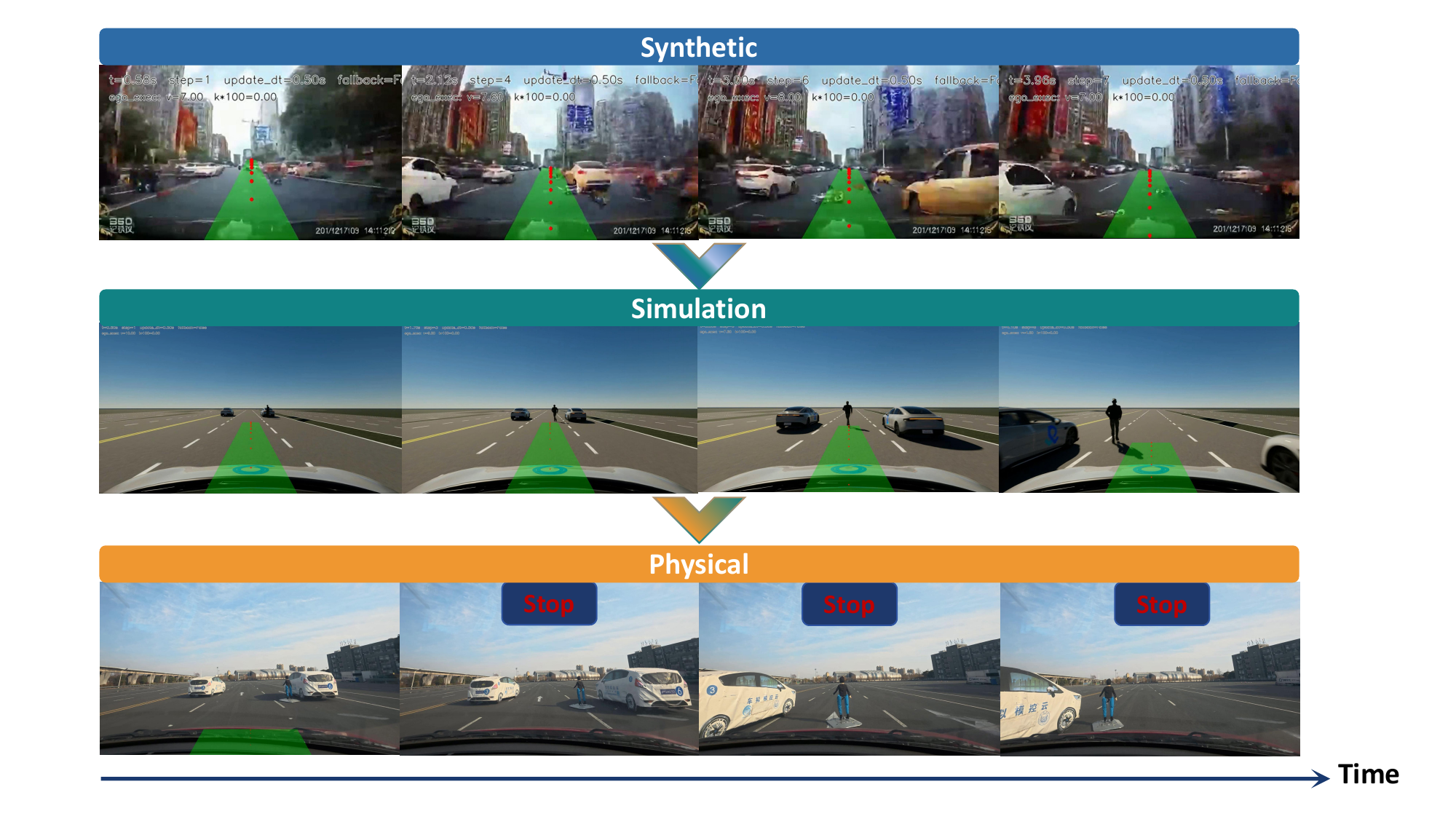}
    \caption{Projected Alpamayo-R1 inference results for the VRU-crossing event across the Synthetic, Simulation, and Physical domains. Predicted trajectories and action responses are overlaid on phase-aligned forward-view frames for qualitative cross-domain comparison.}
    \label{fig:vru-alignment}
\end{figure*}

\begin{figure*}[!t]
    \centering
    \includegraphics[width=\textwidth]{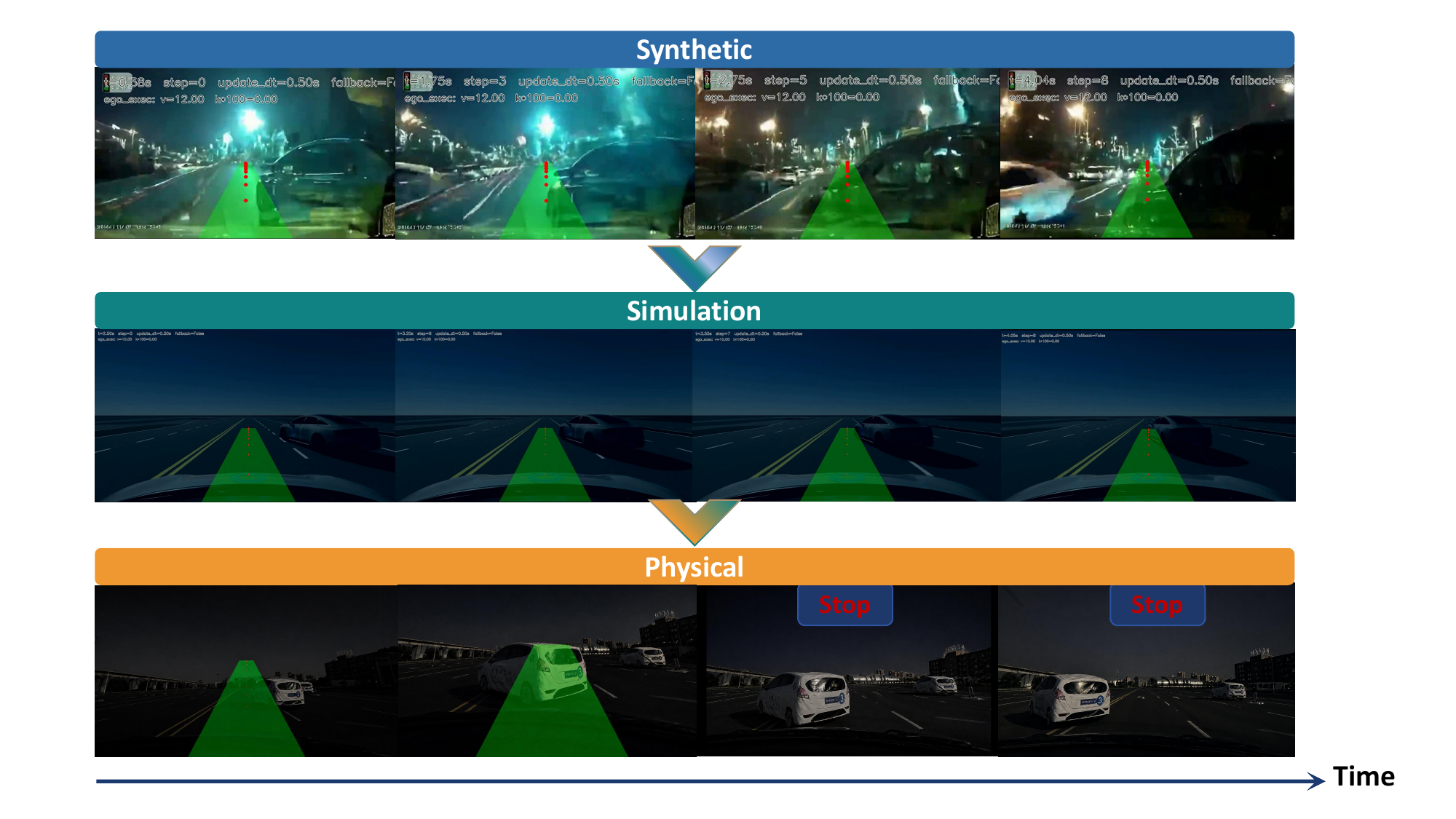}
    \caption{Projected Alpamayo-R1 inference results for the Cut-in event across the Synthetic, Simulation, and Physical domains. Predicted trajectories and action responses are overlaid on phase-aligned forward-view frames for qualitative cross-domain comparison.}
    \label{fig:cutin-alignment}
\end{figure*}

Fig.~\ref{fig:transfer-examples} illustrates the event-matched transfer results and the cross-domain audit information for the two evaluated scenarios. Figs.~\ref{fig:vru-alignment} and ~\ref{fig:cutin-alignment} further provide qualitative examples of Alpamayo-R1 inference across the Synthetic, Simulation, and Physical domains. For each domain, the model's predicted trajectory and action response are projected onto phase-aligned forward-view frames, allowing the cross-domain differences in its reasoning--action behavior to be inspected visually. All trajectory regions, action labels, and audit annotations shown in Figs.~\ref{fig:transfer-examples}--\ref{fig:cutin-alignment} are post-processing visualizations and are not included in the video inputs provided to the VLA models.

The visualizations in Figs.~\ref{fig:vru-alignment} and ~\ref{fig:cutin-alignment} are intended as qualitative complements to the quantitative metrics reported below. Because the same model is evaluated under an event-matched protocol, differences
in the projected trajectory and action response can be inspected against changes in the observation domain while the underlying interaction semantics are preserved. These examples are not used for scoring; all quantitative results are computed from the normalized model outputs using the definitions in Section~\ref{sec:metrics}.

\subsection{Unified VLA metrics and mathematical definitions}
\label{sec:metrics}

Following the interface normalization and evidence extraction described in Section~\ref{subsec:common-representation}, this section defines the scoring rules applied to the resulting common behavior records. Unless otherwise stated, temporal matching, action discretization, and trajectory-feasibility thresholds were fixed before aggregating model results and were applied identically to every scenario, domain, and model.

\subsubsection{Notation, validity, and time alignment}
\label{subsubsec:validity}

For one scenario--domain--model cell, let \(i=1,\ldots,N\) index aligned records, with \(N=9\) in this experiment. Let \(v_i^{\mathrm{txt}}\in\{0,1\}\) indicate nonempty text and \(v_i^{\mathrm{trj}}\in\{0,1\}\) indicate a model-specific trajectory that passes format checks and covers the common interval. The Joint Output Validity Rate (JOVR) is
\begin{equation}
\mathrm{JOVR}=\frac{1}{N}\sum_{i=1}^{N}v_i^{\mathrm{txt}}v_i^{\mathrm{trj}}.
\label{eq:jovr}
\end{equation}
JOVR establishes only whether text and trajectory are jointly evaluable, not whether they are correct.

Label time \(\tau_j\) and prediction time \(t_i\) are paired within \(|t_i-\tau_j|\leq\SI{0.11}{\second}\) by monotone one-to-one dynamic programming. The lexicographic objective first maximizes the number of matches and then minimizes total time error:
\begin{equation}
\mathcal M^{\star}=\argmin_{\mathcal M\in\mathfrak M_{\mathrm{mono}}}
\left[-B|\mathcal M|+\sum_{(i,j)\in\mathcal M}|t_i-\tau_j|\right],
\label{eq:time-matching}
\end{equation}
where \(B\) is sufficiently large and \(\mathfrak M_{\mathrm{mono}}\) is the set of noncrossing matches. Unless first-frame dynamics are explicitly enabled, motion and conflict slots at the earliest label are masked because a single frame does not support a motion trend or conflict evolution.

\subsubsection{Semantic coverage, conditional accuracy, and effective accuracy}
\label{subsubsec:semantics}

Let \(\mathcal Q=\{\mathrm{obj},\mathrm{pos},\mathrm{mot},\mathrm{con}\}\) be the semantic-field set. Indicator \(e_{iq}\) denotes whether slot \(q\) is evaluable in record \(i\), and \(z_{iq}\) denotes whether the model supplies a valid evidence-grounded value. The primary score is \(h_{iq}\in[0,1]\), and \(s_{iq}\in[0,1]\) is a tolerant diagnostic score. For the object set,
\begin{equation}
h_{i,\mathrm{obj}}=s_{i,\mathrm{obj}}=
\frac{2|\widehat O_i\cap O_i|}{|\widehat O_i|+|O_i|}.
\label{eq:object-f1}
\end{equation}
Empty-set behavior follows the evaluator's deterministic convention. Position, motion, and conflict use hard exact-match scores \(h_{iq}=\mathbf 1(\widehat y_{iq}=y_{iq})\). Diagnostic soft scoring assigns 0.5 to adjacent position classes (e.g., \textsc{front} versus \textsc{front-left/right}, or \textsc{left/right} versus the corresponding front-side class) and to adjacent conflict states (\textsc{potential} versus \textsc{active}, and \textsc{no-conflict} versus \textsc{resolved}); all other nonexact matches receive zero.

The three semantic aggregates are
\begin{align}
\mathrm{SemCov}
&=\frac{\sum_{i,q}e_{iq}z_{iq}}{\sum_{i,q}e_{iq}},
\label{eq:semcov}\\
\mathrm{SemAcc}_{\mathrm{cond}}
&=\frac{\sum_{i,q}e_{iq}z_{iq}h_{iq}}{\sum_{i,q}e_{iq}z_{iq}},
\label{eq:semacc-cond}\\
\mathrm{SemAcc}_{\mathrm{eff}}
&=\frac{\sum_{i,q}e_{iq}z_{iq}h_{iq}}{\sum_{i,q}e_{iq}}.
\label{eq:semacc-eff}
\end{align}
Conditional accuracy diagnoses correctness among explicitly answered slots, whereas effective accuracy assigns zero to missing slots and prevents selective silence from inflating performance. Main tables and IVCS use \(\mathrm{SemAcc}_{\mathrm{eff}}\). Soft accuracy replaces \(h_{iq}\) with \(s_{iq}\) and remains diagnostic only.

\subsubsection{Critical-interaction recognition and explicit action}
\label{subsubsec:interaction-action}

The critical-interaction tuple comprises object type, relative position, and motion pattern, excluding conflict phase. Define
\begin{equation}
E_i^{\mathrm{cia}}=e_{i,\mathrm{obj}}e_{i,\mathrm{pos}}e_{i,\mathrm{mot}},\qquad
C_i^{\mathrm{cia}}=\prod_{q\in\{\mathrm{obj},\mathrm{pos},\mathrm{mot}\}}
z_{iq}\,\mathbf 1(h_{iq}=1).
\label{eq:cia-indicators}
\end{equation}
The effective Critical Interaction Accuracy is
\begin{equation}
\mathrm{CIA}_{\mathrm{eff}}=
\frac{\sum_i E_i^{\mathrm{cia}}C_i^{\mathrm{cia}}}{\sum_i E_i^{\mathrm{cia}}}.
\label{eq:cia-eff}
\end{equation}
The object set must attain an F1 score of one, and position and motion must both be exact; CIA is thus stricter than field-wise semantic accuracy.

Let \(Y_i^{\mathrm{lon}}\) and \(Y_i^{\mathrm{lat}}\) be the allowable longitudinal and lateral action sets, \(\widehat a_i^{\mathrm{lon}}\) and \(\widehat a_i^{\mathrm{lat}}\) the extracted actions, and \(z_i^{\mathrm{lon}},z_i^{\mathrm{lat}}\) their evidence-validity indicators. Joint effective Action Accuracy is
\begin{equation}
\mathrm{ActAcc}_{\mathrm{eff}}=\frac{1}{N}\sum_i
z_i^{\mathrm{lon}}z_i^{\mathrm{lat}}
\mathbf 1\!\left(\widehat a_i^{\mathrm{lon}}\in Y_i^{\mathrm{lon}}\right)
\mathbf 1\!\left(\widehat a_i^{\mathrm{lat}}\in Y_i^{\mathrm{lat}}\right).
\label{eq:action-accuracy}
\end{equation}
This quantity evaluates the textual action commitment, not whether the trajectory implements it.

\subsubsection{Common trajectory, inferred action, and geometric interpretation}
\label{subsubsec:trajectory-actions}

Each trajectory is interpolated at \(t_0=0\), \(t_1=0.5\), and \(t_2=1.0\)~s. Let \(\overline{\bm p}_{i,k}=(s_{i,k},l_{i,k})\), where \(s\) points forward and \(l\) is positive to the left. With \(\Delta t=\SI{0.5}{\second}\),
\begin{equation}
\begin{aligned}
v_{i,1}^{s}&=\frac{s_{i,1}-s_{i,0}}{\Delta t}, &
v_{i,2}^{s}&=\frac{s_{i,2}-s_{i,1}}{\Delta t},\\
a_i^{s}&=\frac{v_{i,2}^{s}-v_{i,1}^{s}}{\Delta t}, &
\Delta l_i&=l_{i,2}-l_{i,0}.
\end{aligned}
\label{eq:trajectory-kinematics}
\end{equation}
The trajectory-implied longitudinal action is
\begin{equation}
\widehat a_i^{\mathrm{trj,lon}}=
\begin{cases}
\textsc{Stop}, & v_{i,2}^{s}\leq\SI{0.5}{\meter\per\second},\\
\textsc{Decelerate}, & a_i^{s}\leq-\SI{0.5}{\meter\per\second\squared},\\
\textsc{Accelerate}, & a_i^{s}\geq\SI{0.5}{\meter\per\second\squared},\\
\textsc{Maintain}, & \text{otherwise},
\end{cases}
\label{eq:longitudinal-action}
\end{equation}
and the lateral action is
\begin{equation}
\widehat a_i^{\mathrm{trj,lat}}=
\begin{cases}
\textsc{Left}, & \Delta l_i\geq\SI{0.3}{\meter},\\
\textsc{Right}, & \Delta l_i\leq-\SI{0.3}{\meter},\\
\textsc{Straight}, & \text{otherwise}.
\end{cases}
\label{eq:lateral-action}
\end{equation}
The final-segment speed separates stopping from motion, mean acceleration separates acceleration from deceleration, and net lateral displacement represents the \SI{1}{\second} lane-direction response. These are evaluator thresholds, not a full vehicle-dynamics controller.

\subsubsection{Feasibility, trajectory quality, and text--trajectory consistency}
\label{subsubsec:trajectory-quality}

Let \(\Delta\bm p_{i,k}=\overline{\bm p}_{i,k}-\overline{\bm p}_{i,k-1}\), \(u_{i,k}=\|\Delta\bm p_{i,k}\|_2/\Delta t\), and \(v_{i,k}^{l}=\Delta l_{i,k}/\Delta t\). The feasibility indicator is
\begin{equation}
\begin{aligned}
F_i={}&\prod_{k=1}^{2}
\mathbf 1(\Delta s_{i,k}\geq-0.2)
\mathbf 1(0\leq u_{i,k}\leq25)\\
&\times\mathbf 1(|v_{i,k}^{l}|\leq5)
\mathbf 1(|a_i^{s}|\leq8),
\end{aligned}
\label{eq:feasibility}
\end{equation}
where the associated units are meters, meters per second, and meters per second squared. These limits reject obvious reverse motion, excessive speed or acceleration, and lateral jumps; they are not certification limits. Let \(R_i^{\mathrm{lon}}\) and \(R_i^{\mathrm{lat}}\) indicate that the inferred actions fall in the allowable sets, and \(R_i^{\mathrm{trj}}=R_i^{\mathrm{lon}}R_i^{\mathrm{lat}}\). Then
\begin{equation}
\mathrm{STFR}=\frac{1}{N}\sum_i v_i^{\mathrm{trj}}F_i,\qquad
\mathrm{TQSR}=\frac{1}{N}\sum_i v_i^{\mathrm{trj}}F_iR_i^{\mathrm{trj}}.
\label{eq:trajectory-rates}
\end{equation}
The Spatiotemporal Feasibility Rate (STFR) requires only a valid feasible trajectory; TQSR additionally requires correct longitudinal and lateral responses.

Action--Trajectory Consistency (ATC) ignores the ground-truth label and measures agreement between the two model outputs:
\begin{equation}
\mathrm{ATC}_{\mathrm{eff}}=\frac{1}{N}\sum_i v_i^{\mathrm{trj}}
z_i^{\mathrm{lon}}z_i^{\mathrm{lat}}
\mathbf 1\!\left(\widehat a_i^{\mathrm{lon}}=\widehat a_i^{\mathrm{trj,lon}}\right)
\mathbf 1\!\left(\widehat a_i^{\mathrm{lat}}=\widehat a_i^{\mathrm{trj,lat}}\right).
\label{eq:atc}
\end{equation}
A high ATC is not necessarily safe because text and trajectory may agree on the same incorrect action. Conversely, a high TQSR and low ATC indicate a label-consistent trajectory whose textual explanation does not describe that trajectory.

\subsubsection{Sustained longitudinal risk response and latency}
\label{subsubsec:risk-response}

Risk onset \(t_r\) is taken from the top-level \texttt{risk\_time\_s} label when available, otherwise from the first \texttt{risk\_visible=true} label. A candidate also requires \(t_i\geq t_r\) and a \textsc{potential} or \textsc{active} conflict phase. By default, \(K=2\) consecutive candidates must be no more than \SI{0.6}{\second} apart, imply \textsc{decelerate} or \textsc{stop}, and satisfy the longitudinal label. Define
\begin{equation}
I_i^{\mathrm{def}}=
\mathbf 1\!\left(\widehat a_i^{\mathrm{trj,lon}}
\in\{\textsc{Decelerate},\textsc{Stop}\}\right)R_i^{\mathrm{lon}}.
\label{eq:defensive-indicator}
\end{equation}
The sustained-response time is then
\begin{equation}
\begin{aligned}
t_{\mathrm{resp}}=\min_i\{t_i:\;&
\textstyle\prod_{j=0}^{K-1}I_{i+j}^{\mathrm{def}}=1,\\
&0<t_{i+j+1}-t_{i+j}\leq0.6,\;
j=0,\ldots,K-2\}.
\end{aligned}
\label{eq:response-time}
\end{equation}
If the set is nonempty, \(\mathrm{RSR}=1\) and \(\mathrm{RL}=\max(0,t_{\mathrm{resp}}-t_r)\); otherwise \(\mathrm{RSR}=0\) and latency is missing rather than assigned an arbitrary large value. Tables use S@x.xx for success and latency in seconds, and F for failure. RSR measures sustained longitudinal deceleration or stopping only; lateral avoidance remains represented by TQSR and ATC.

\subsubsection{Integrated score}
\label{subsubsec:ivcs}

The Integrated VLA Capability Score is
\begin{equation}
\begin{aligned}
\mathrm{IVCS}={}&0.10\,\mathrm{JOVR}
+0.30\,\mathrm{SemAcc}_{\mathrm{eff}}\\
&+0.20\,\mathrm{CIA}_{\mathrm{eff}}
+0.30\,\mathrm{TQSR}
+0.10\,\mathrm{RSR}.
\end{aligned}
\label{eq:ivcs}
\end{equation}
The weighting assigns 80\% of the score to effective semantics, critical-interaction recognition, and trajectory quality, and 10\% each to interface validity and event-level risk response. ActAcc and ATC remain diagnostics to avoid double-counting semantic and trajectory behavior. IVCS does not constitute a vehicle-safety certification score.

\subsection{Complete result matrix}
\label{subsec:complete-results}

Table~\ref{tab:complete-results} reports the results for all 18 scenario--domain--model cells. Under the strict joint criterion, CIA is zero in most cells, showing that simultaneously binding object identity, relative position, and motion remains one of the most difficult semantic requirements. High JOVR does not substitute for correct content.

\begin{table*}[!t]
\centering
\caption{Complete cross-domain evaluation results for all scenario--domain--model cells.}
\label{tab:complete-results}
\scriptsize
\setlength{\tabcolsep}{3.2pt}
\begin{adjustbox}{max width=\textwidth}
\begin{tabular}{lllccccccc}
\toprule
Scenario & Domain & Model & {JOVR} & {\makecell{SemAcc\\\(_{\mathrm{eff}}\)}} & {\makecell{CIA\\\(_{\mathrm{eff}}\)}} & {\makecell{ActAcc\\\(_{\mathrm{eff}}\)}} & {TQSR} & {\makecell{ATC\\\(_{\mathrm{eff}}\)}} & {Risk response} \\
\midrule
Cut-in & Synthetic  & OpenEMMA    & 1.000 & 0.294 & 0.000 & 0.444 & 0.444 & 1.000 & F \\
Cut-in & Synthetic  & LLaViDA     & 0.889 & 0.000 & 0.000 & 0.333 & 0.111 & 0.556 & S@1.50 \\
Cut-in & Synthetic  & Alpamayo-R1 & 1.000 & 0.265 & 0.000 & 0.000 & 0.444 & 0.000 & F \\
Cut-in & Simulation & OpenEMMA    & 1.000 & 0.294 & 0.000 & 0.444 & 0.444 & 1.000 & F \\
Cut-in & Simulation & LLaViDA     & 0.667 & 0.000 & 0.000 & 0.111 & 0.111 & 0.222 & F \\
Cut-in & Simulation & Alpamayo-R1 & 1.000 & 0.324 & 0.000 & 0.000 & 1.000 & 0.000 & S@0.00 \\
Cut-in & Physical   & OpenEMMA    & 1.000 & 0.324 & 0.000 & 0.556 & 0.444 & 0.778 & S@2.00 \\
Cut-in & Physical   & LLaViDA     & 0.556 & 0.000 & 0.000 & 0.333 & 0.000 & 0.444 & F \\
Cut-in & Physical   & Alpamayo-R1 & 1.000 & 0.353 & 0.000 & 0.000 & 0.444 & 0.000 & F \\
\midrule
VRU & Synthetic  & OpenEMMA    & 1.000 & 0.000 & 0.000 & 0.222 & 0.111 & 0.222 & F \\
VRU & Synthetic  & LLaViDA     & 0.111 & 0.000 & 0.000 & 0.000 & 0.000 & 0.000 & F \\
VRU & Synthetic  & Alpamayo-R1 & 1.000 & 0.176 & 0.000 & 0.000 & 1.000 & 0.000 & S@0.00 \\
VRU & Simulation & OpenEMMA    & 1.000 & 0.294 & 0.000 & 0.889 & 0.333 & 0.222 & F \\
VRU & Simulation & LLaViDA     & 0.667 & 0.000 & 0.000 & 0.222 & 0.222 & 0.222 & F \\
VRU & Simulation & Alpamayo-R1 & 1.000 & 0.382 & 0.125 & 0.000 & 0.222 & 0.000 & F \\
VRU & Physical   & OpenEMMA    & 1.000 & 0.216 & 0.000 & 0.556 & 0.889 & 0.333 & S@0.00 \\
VRU & Physical   & LLaViDA     & 1.000 & 0.000 & 0.000 & 0.444 & 0.222 & 0.333 & S@0.00 \\
VRU & Physical   & Alpamayo-R1 & 1.000 & 0.441 & 0.125 & 0.000 & 0.222 & 0.000 & F \\
\bottomrule
\end{tabular}
\end{adjustbox}
\end{table*}

\begin{figure*}[!t]
    \centering
    \includegraphics[width=\textwidth]{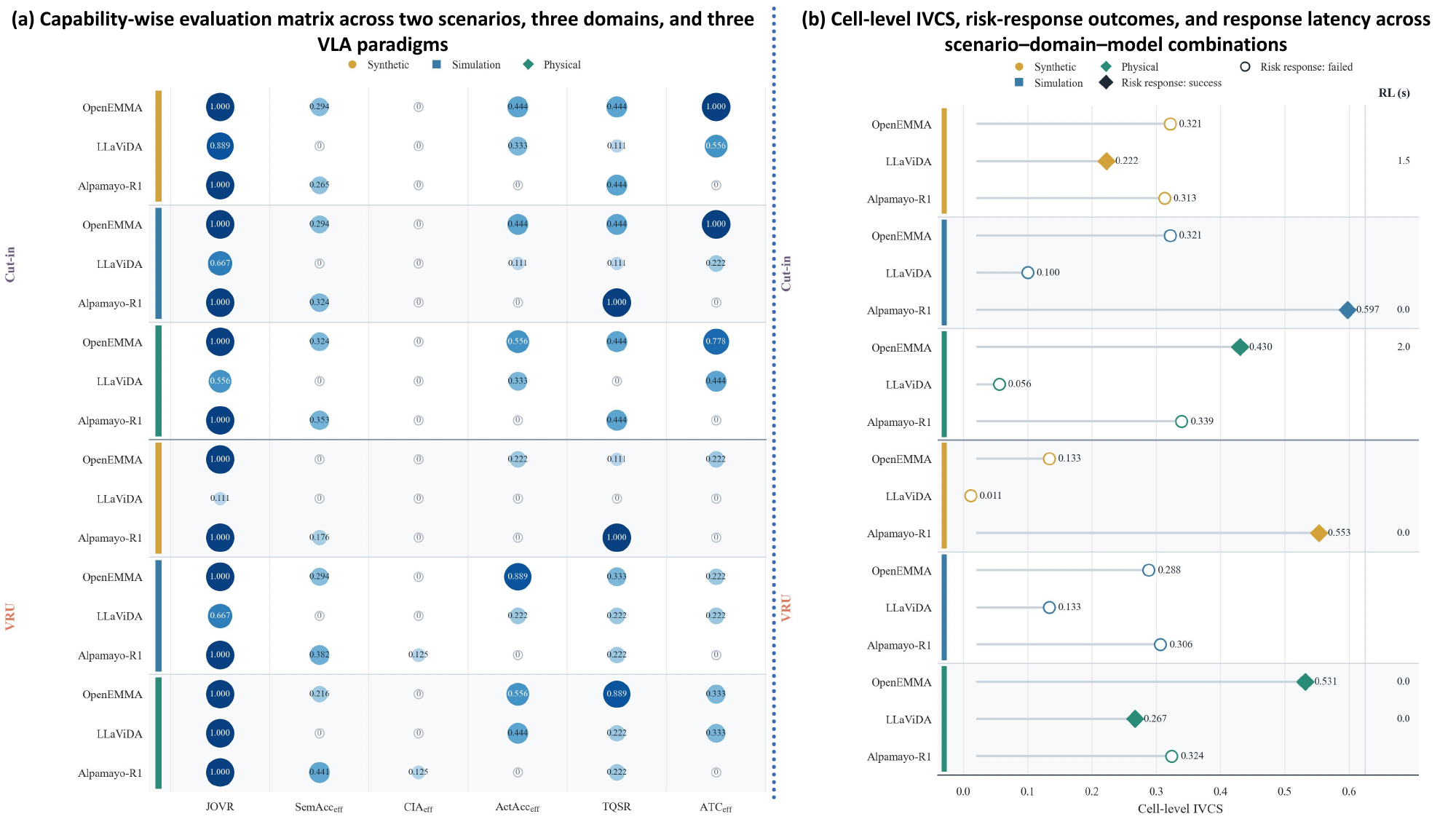}
    \caption{(a) Capability-wise evaluation matrix across two scenarios, three domains, and three VLA paradigms. (b) Cell-level IVCS, risk-response outcomes, and response latency across scenario--domain--model combinations.}
    \label{fig:results-matrix}
\end{figure*}

Fig.~\ref{fig:results-matrix} uses numerical and area encoding: circle area and lightness jointly show component scores, while the right-hand panel reports risk-response state, latency, and cell-level IVCS computed from the reported components. The design makes zero-valued bottlenecks, scenario/domain groupings, and capability differences visible without allowing a large dark tile to dominate the evidence.

\subsection{RQ1: domain-conditioned VLA performance is scenario dependent}
\label{subsec:rq1}

The domain macro averages in Table~\ref{tab:domain-results} rank Physical, Simulation, and Synthetic at IVCS values of 0.325, 0.291, and 0.259. Physical exceeds Simulation by 11.5\% and Synthetic by 25.3\%, while Simulation exceeds Synthetic by 12.4\% in the reported aggregation. The Physical-domain advantage comes mainly from JOVR, \(\mathrm{SemAcc}_{\mathrm{eff}}\), and RSR rather than TQSR: Simulation attains a TQSR of 0.389, slightly above the Physical value of 0.370. Thus, greater nominal realism does not imply superiority on every component.

\begin{table*}[!t]
\centering
\caption{Macro-averaged results by domain.}
\label{tab:domain-results}
\small
\setlength{\tabcolsep}{6.8pt}
\begin{tabular}{lccccccc}
\toprule
Domain & {JOVR} & {\(\mathrm{SemAcc}_{\mathrm{eff}}\)} & {\(\mathrm{CIA}_{\mathrm{eff}}\)} & {TQSR} & {RSR} & {IVCS} & Rank \\
\midrule
Physical   & 0.926 & 0.222 & 0.021 & 0.370 & 0.500 & 0.325 & 1 \\
Simulation & 0.889 & 0.216 & 0.021 & 0.389 & 0.167 & 0.291 & 2 \\
Synthetic  & 0.833 & 0.122 & 0.000 & 0.352 & 0.333 & 0.259 & 3 \\
\bottomrule
\end{tabular}
\end{table*}

Scenario stratification reveals a rank reversal. For Cut-in, the Physical/Simulation/Synthetic IVCSs are 0.275/0.340/0.285, so Simulation ranks first. For VRU crossing, they are 0.374/0.243/0.232, so Physical ranks first. The macro ordering Physical \(>\) Simulation \(>\) Synthetic describes model performance for the present scenario and model sets; it is neither a context-free ranking of data value nor a rule for arbitrary events.

\begin{figure*}[!t]
    \centering
    \includegraphics[width=\textwidth]{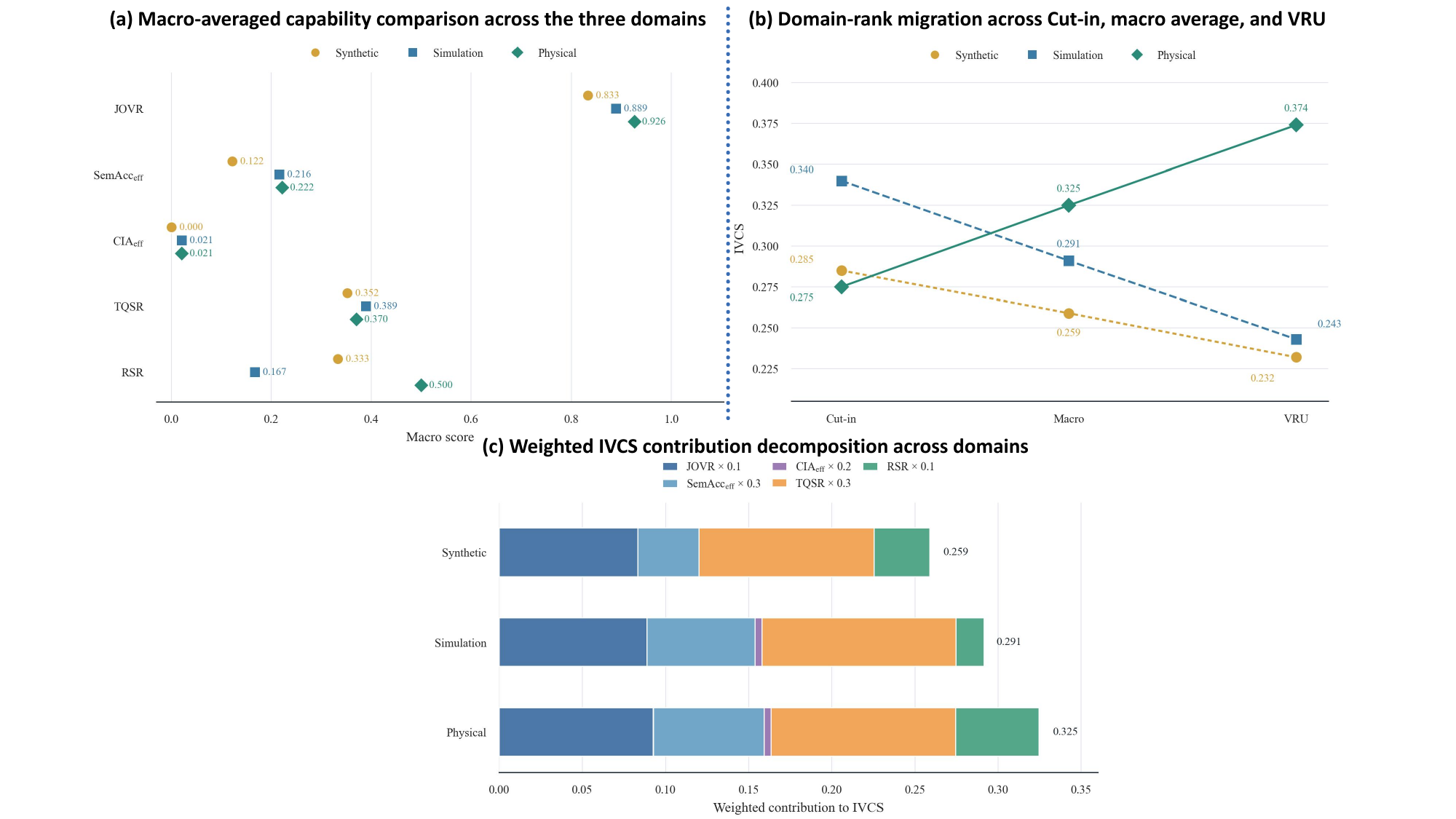}
    \caption{(a) Macro-averaged capability comparison across the three domains. (b) Domain-rank migration across Cut-in, macro average, and VRU. (c) Weighted IVCS contribution decomposition across domains.}
    \label{fig:domain-analysis}
\end{figure*}

Fig.~\ref{fig:domain-analysis} combines three views: a Cleveland dot plot for the five capability scores, a Cut-in--macro--VRU slope plot that exposes domain-rank migration, and stacked contributions to IVCS. The figure therefore shows both the aggregate ordering and why and when it fails.

\subsection{RQ2: VLA differences arise from capability-chain balance}
\label{subsec:rq2}

Table~\ref{tab:paradigm-results} shows IVCSs of 0.405, 0.338, and 0.131 for Alpamayo-R1, OpenEMMA, and LLaViDA. Alpamayo-R1 exceeds OpenEMMA by 20.0\% and LLaViDA by 208.3\% in the reported aggregation. OpenEMMA and Alpamayo-R1 both attain JOVR=1.000, so their difference is not explained by the ability to return an output. Alpamayo-R1 combines higher effective semantic accuracy (0.324 versus 0.237), nonzero CIA (0.042 versus 0), and higher TQSR (0.555 versus 0.444). This result is consistent with its intended reasoning--action coupling \cite{nvidia2025alpamayo}; because internal variables are not observed, it is not a causal attribution to a specific mechanism.

\begin{table*}[!t]
\centering
\caption{Macro-averaged results by VLA paradigm.}
\label{tab:paradigm-results}
\small
\setlength{\tabcolsep}{6.8pt}
\begin{tabular}{lccccccc}
\toprule
Model & {JOVR} & {\(\mathrm{SemAcc}_{\mathrm{eff}}\)} & {\(\mathrm{CIA}_{\mathrm{eff}}\)} & {TQSR} & {RSR} & {IVCS} & Rank \\
\midrule
Alpamayo-R1 & 1.000 & 0.324 & 0.042 & 0.555 & 0.333 & 0.405 & 1 \\
OpenEMMA    & 1.000 & 0.237 & 0.000 & 0.444 & 0.333 & 0.338 & 2 \\
LLaViDA     & 0.648 & 0.000 & 0.000 & 0.111 & 0.333 & 0.131 & 3 \\
\bottomrule
\end{tabular}
\end{table*}

LLaViDA's low IVCS combines interface absence (JOVR=0.648), missing or incorrect evidence slots (\(\mathrm{SemAcc}_{\mathrm{eff}}=0\)), and insufficient trajectory quality (TQSR=0.111). This finding does not contradict the model's reported results under its native training and benchmark \cite{liu2025llavida}; it shows that cross-model reproduction depends on prompt adaptation, coordinates, trajectory parsing, and the common window. The ranking remains the same in both scenarios: the Cut-in/VRU IVCSs are 0.416/0.394 for Alpamayo-R1, 0.358/0.318 for OpenEMMA, and 0.126/0.137 for LLaViDA.

\begin{figure*}[!t]
    \centering
    \includegraphics[width=\textwidth]{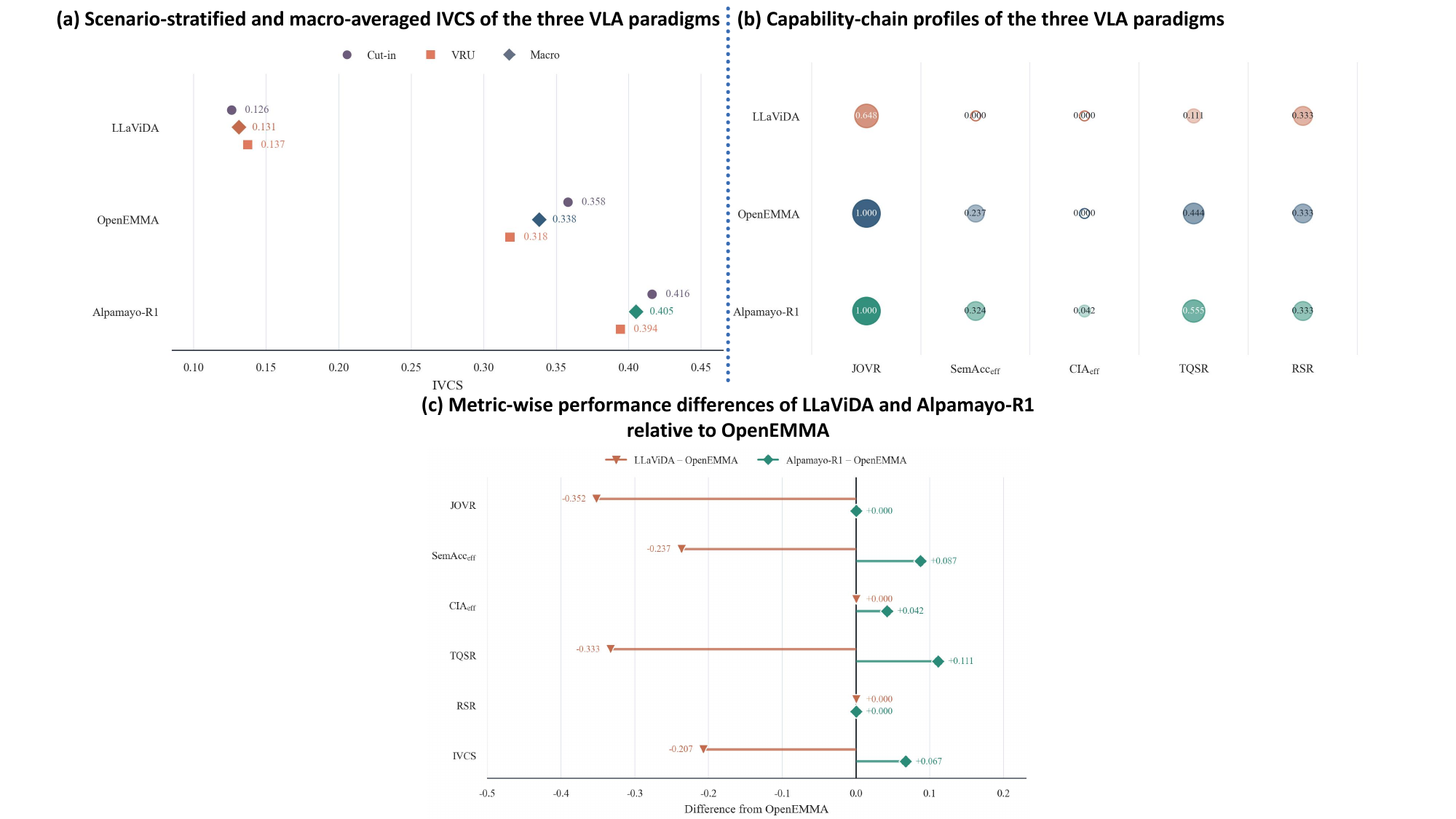}
    \caption{(a) Scenario-stratified and macro-averaged IVCS of the three VLA paradigms. (b) Capability-chain profiles of the three VLA paradigms. (c) Metric-wise performance differences of LLaViDA and Alpamayo-R1 relative to OpenEMMA.}
    \label{fig:model-analysis}
\end{figure*}

Fig.~\ref{fig:model-analysis} presents scenario ranges with a macro diamond, capability profiles, and differences from OpenEMMA. These views separate overall level, cross-scenario variation, and the sign of each capability change.

\subsection{RQ3: foundation-model scale and MoE architecture under a common interface}
\label{subsec:rq3}

To reduce interface confounding, the controlled RQ3 comparison fixes the OpenEMMA-style data flow, task definition, and parser while varying Qwen3-VL Dense/MoE configurations \cite{bai2025qwen3vl} and Seed1.6 Flash \cite{bytedance2025seed16}. Alpamayo-R1 retains its native driving-specific reasoning and trajectory interface and is included only as an external driving-specific reference. It is not used to infer causal effects of Dense/MoE architecture or foundation-model scale. For MoE systems, Total/Active denotes total parameters and approximate per-token active parameters; it is not a measurement of FLOPs, end-to-end latency, or memory.

\begin{table*}[!t]
\centering
\caption{Comparison of foundation-model scale and architecture.}
\label{tab:architecture-results}
\scriptsize
\setlength{\tabcolsep}{3.5pt}
\begin{adjustbox}{max width=\textwidth}
\begin{tabular}{lllcccccccc}
\toprule
Model & Architecture & {Total/Active (B)} & {JOVR} & {SemCov} & {\makecell{SemAcc\\\(_{\mathrm{eff}}\)}} & {\makecell{CIA\\\(_{\mathrm{eff}}\)}} & {TQSR} & {RSR} & {IVCS} & {vs. 8B} \\
\midrule
Qwen3-VL-8B        & Dense            & 8/8    & 1.000 & 0.642 & 0.273 & 0.000 & 0.056 & 0.667 & 0.265 & Reference \\
Qwen3-VL-32B       & Dense            & 32/32  & 1.000 & 0.618 & 0.240 & 0.000 & 0.426 & 0.667 & 0.366 & +38.1\% \\
Qwen3-VL-30B-A3B   & MoE              & 30/3   & 0.944 & 0.569 & 0.217 & 0.000 & 0.630 & 0.500 & 0.398 & +50.2\% \\
Qwen3-VL-235B-A22B & MoE              & 235/22 & 1.000 & 0.662 & 0.260 & 0.000 & 0.426 & 0.500 & 0.356 & +34.1\% \\
Seed1.6 Flash      & Proprietary      & --     & 1.000 & 0.593 & 0.237 & 0.000 & 0.444 & 0.333 & 0.338 & +27.2\% \\
Alpamayo-R1        & Driving-specific & --     & 1.000 & 0.534 & 0.324 & 0.042 & 0.555 & 0.333 & 0.405 & +52.8\% \\
\bottomrule
\end{tabular}
\end{adjustbox}
\end{table*}

From 8B to 32B Dense, TQSR rises from 0.056 to 0.426 and IVCS from 0.265 to 0.366, while \(\mathrm{SemAcc}_{\mathrm{eff}}\) decreases from 0.273 to 0.240. Capacity growth therefore does not improve every component simultaneously. Qwen3-VL-30B-A3B activates approximately 3B parameters and attains the highest Qwen TQSR (0.630) and IVCS (0.398). Relative to 32B Dense, it trades lower JOVR, semantic coverage, and RSR for a large trajectory gain. Qwen3-VL-235B-A22B has the greatest semantic coverage (0.662), but its TQSR (0.426) and IVCS (0.356) do not exceed those of the smaller MoE. The current experiment therefore does not support a monotone relation between total parameters and integrated driving capability.

\begin{figure*}[!t]
    \centering
    \includegraphics[width=\textwidth]{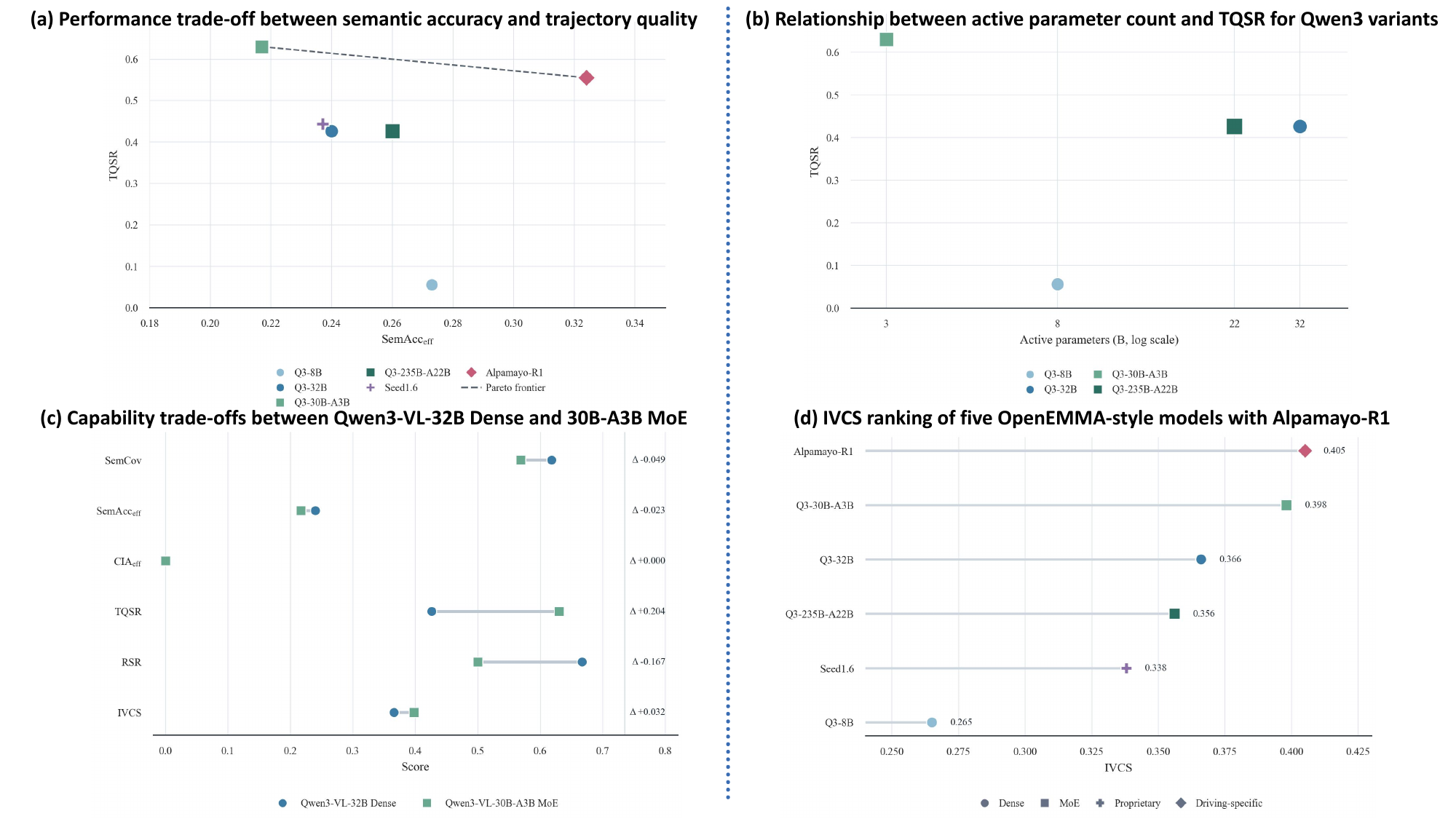}
    \caption{(a) Performance trade-off between semantic accuracy and trajectory quality. (b) Relationship between active parameter count and TQSR for Qwen3 variants. (c) Capability trade-offs between Qwen3-VL-32B Dense and 30B-A3B MoE. (d) IVCS ranking of five OpenEMMA-style models with Alpamayo-R1.}
    \label{fig:scaling-analysis}
\end{figure*}

Alpamayo-R1 obtains the highest IVCS in Table~\ref{tab:architecture-results} (0.405) by combining effective semantics, the only nonzero CIA, and a high TQSR. Together with Qwen3-VL-30B-A3B, it occupies the leading semantic--trajectory region in Fig.~\ref{fig:scaling-analysis}. The protocol-level evidence therefore indicates that driving-specific alignment and behavior-chain coordination can be more effective than increasing the total size of a general-purpose foundation model for the evaluated cases; it does not imply a universal scaling law.

Fig.~\ref{fig:scaling-analysis} avoids placing systems with unknown or incomparable parameter definitions on a single size axis. Its Pareto panel shows semantic--trajectory coordination; the active-parameter panel uses a log axis only for comparable Qwen variants; the Dense--MoE dumbbell plot shows capability trade-offs; and the final panel ranks the six configurations by IVCS.

\section{Discussion}
\label{sec:discussion}

\subsection{The central value of SSP: three domains as one event evidence chain}
\label{subsec:discussion-chain}

The central methodological change in SSP is not a new video-to-simulator or video-to-track generation algorithm, but a different unit of comparison for cross-domain VLA evaluation. Conventional cross-domain comparisons often search each domain for a nominally similar case; road geometry, participants, risk phase, and normative response may then change implicitly, leaving no unique attribution for the score difference. SSP begins with a specific synthetic event, extracts its executable semantics, and constructs simulation and physical tasks from a shared topology, role assignment, conflict anchor, and phase definition. Cross-domain inputs therefore share an auditable event source.

This is a system-level contribution rather than a claim that any single representation or generator is replaced. The interaction graph packages and validates road--participant--event information; the methodological chain is source anchoring, structural reconstruction in simulation, physical execution confirmation, and unified behavior evaluation. Relative to our earlier work, the research object and output change: an extensible synthetic long-tail source replaces a real interaction video as input, while three-domain matched videos and their VLA evidence replace a single proving-ground test case. The question changes from how to generate a realistic or adversarial physical test to what kind of evidence each domain contributes to VLA evaluation.

The Syn2Sim2Phy order reflects increasing implementation cost, engineering constraint, and completeness of the execution chain. Synthetic data expand candidate long-tail events; Simulation tests topology, triggering, and participant executability and supports low-cost diagnosis; Physical data introduce real imaging, vehicle motion, and control error for confirmation. This order does not imply that a later domain must score higher. It defines an engineering funnel: broaden coverage in a low-cost domain, reduce uncertainty in a controllable domain, and reserve limited physical resources for high-value confirmation.

\subsection{Domain-conditioned performance has an aggregate trend but strong scenario dependence}
\label{subsec:discussion-domain}

RQ1 yields a macro IVCS order of Physical (0.325) \(>\) Simulation (0.291) \(>\) Synthetic (0.259). On the current two scenarios and three VLA paradigms, Physical inputs support a more balanced combination of output validity, semantic evidence, and sustained risk response. The advantage is not shared by every component: Simulation has a TQSR of 0.389 versus 0.370 for Physical, and Synthetic has an RSR of 0.333 versus 0.167 for Simulation. The macro average cannot therefore be restated as ``more realistic data always make the model better.''

The scenario-level reversal clarifies this qualification. One plausible interpretation is that Cut-in recognition depends strongly on vehicle contour, adjacent-lane structure, and lateral intrusion, for which regularized lane boundaries and stable target motion in Simulation may provide unusually clear evidence. VRU crossing may depend more on small-target class, real texture, and continuous lateral motion, for which the Physical clip provides richer cues. This is an inference from the observed pattern, not a controlled causal decomposition. Different events require different visual evidence, and a single realism variable cannot summarize the interactions among training distribution, target scale, motion cue, and execution noise.

Accordingly, the domains should be treated as complementary test functions rather than a value hierarchy inferred from IVCS. Synthetic data suit large-scale screening, prompt-sensitivity analysis, and exposure of obvious failures. Simulation suits controlled repetition, parameter sweeps, and local trajectory diagnosis. Physical data suit confirmation under real imaging, motion, and execution disturbance. Their roles follow from testing properties rather than from a single aggregate ranking.

\subsection{VLA paradigm differences reflect multi-stage coordination}
\label{subsec:discussion-paradigm}

RQ2 gives IVCSs of 0.405, 0.338, and 0.131 for Alpamayo-R1, OpenEMMA, and LLaViDA. Since Alpamayo-R1 and OpenEMMA both have JOVR=1.000, their difference cannot be assigned to output completeness. Alpamayo-R1 improves \(\mathrm{SemAcc}_{\mathrm{eff}}\), CIA, and TQSR together: it more often provides both an explicit participant relationship and a label-consistent short-horizon response. The pattern is consistent with the model's stated focus on long-tail reasoning and continuous action \cite{nvidia2025alpamayo}, but its training data, post-training, and action decoder are not controlled here, so the result cannot be uniquely attributed to causal reasoning or a diffusion decoder.

OpenEMMA illustrates the dual nature of an open interface. It returns complete outputs in every cell and supports replacement of the underlying VLM at relatively low adapter cost \cite{xing2025openemma}, yet JOVR=1 does not produce high CIA. Format stability and interaction understanding are distinct capabilities. LLaViDA exhibits missing outputs, zero effective semantic evidence, and low trajectory quality in this implementation. This does not invalidate its native benchmark results \cite{liu2025llavida}; it demonstrates that prompts, coordinate conventions, text-to-trajectory parsing, and context windows are part of a cross-paper comparison and must be disclosed.

The near-zero CIA in the 18-cell matrix is a more general finding than the rank order. CIA requires the correct participant, relative position, and motion pattern to be bound at the same time. Improving field-average semantics is insufficient if ``who,'' ``where,'' and ``how moving'' refer to different entities or instants. VLA development should therefore treat entity binding, temporal tracking, and action conditioning as a joint target rather than merely extending natural-language explanations.

\subsection{Scale and MoE: active efficiency does not replace capability balance}
\label{subsec:discussion-scaling}

RQ3 provides a more controlled observation because the OpenEMMA-style interface is fixed. Scaling Qwen3-VL from 8B to 32B Dense accompanies an increase in TQSR from 0.056 to 0.426 and IVCS from 0.265 to 0.366, but \(\mathrm{SemAcc}_{\mathrm{eff}}\) decreases from 0.273 to 0.240. Additional capacity does not move all metrics in the same direction.

Qwen3-VL-30B-A3B attains a TQSR of 0.630 and an IVCS of 0.398 with approximately 3B active parameters, leading the Qwen variants. Compared with 32B Dense, it gains trajectory quality while losing some JOVR, semantic coverage, and RSR. The larger 235B-A22B has the highest semantic coverage but does not exceed the smaller MoE in TQSR or IVCS. These results frame MoE value as selective activation that can improve effective capacity, not as a monotone law relating total parameter count to driving capability. Routing, allocation of visual tokens, prompting, and the action interface may matter as much as total or active parameter count.

Alpamayo-R1's IVCS of 0.405 is slightly above Qwen3-VL-30B-A3B and has a more balanced semantic--trajectory profile. The limited engineering implication is that driving-specific data and reasoning--action alignment may be more effective for safety-critical interaction than scaling a general-purpose VLM alone. A deployment comparison would also require measured latency, throughput, memory, and energy; because those measurements are absent, no real-time-efficiency conclusion is drawn.

\subsection{Implications for automated-driving test systems and benchmark design}
\label{subsec:implications}

SSP supports an operational three-stage test strategy. The Synthetic stage runs broad scenario coverage and filters missing outputs, incorrect critical interactions, and obviously infeasible trajectories. The Simulation stage reproduces candidate failures under fixed seeds and controlled triggers and uses privileged logs to determine whether the failure originates in visual understanding, temporal judgment, or action generation. Only cases that are valuable, executable, and transfer-qualified proceed to the Physical stage, where a real camera and vehicle execution chain test whether the behavior persists. This reduces the cost of testing every event physically while avoiding claims of real-world safety based solely on generated or simulated evidence.

For benchmark design, cross-domain event matching should be stricter than sharing the same scenario label. A future cross-domain VLA dataset should provide not only three videos, but also structured scene knowledge, topology and anchor mappings, event phases, compiler parameters, and execution logs. A leaderboard should expose output validity, effective semantics, critical interaction, trajectory quality, and sustained risk response together with scenario strata. Figs.~\ref{fig:results-matrix}--\ref{fig:scaling-analysis} use a matrix, slope plot, difference plot, and Pareto view precisely to reveal capability structure and counterexamples that a single bar height would conceal.

SSP also restricts the LLM evaluator to evidence extraction. This reduces dependence on style, answer length, and judge preference and makes every slot traceable to the original model output. For a safety-critical system, such provenance is more useful than an unexplained semantic score. Human experts should validate the extractor and label quality, rather than be replaced by an unconstrained LLM grade.

\subsection{Threats to validity and limitations}
\label{subsec:limitations}

\paragraph{Internal validity}
Results are computed from a fixed paired evaluation set under controlled prompt, decoding, preprocessing, and parser configurations. The phase-aligned labels support behavior-chain diagnosis within each asset, and interpretation focuses on protocol-level contrasts and scenario-stratified patterns. The same scene-card and evidence-record structure can accommodate additional scenarios and executions without changing the metric definitions.

\paragraph{Scene-transfer validity}
A shared \(K_s^{\star}\) reduces scenario confounding but cannot keep every unmodeled factor constant. A synthetic source may contain physical inconsistency, CARLA behavior may change motion details, and proving-ground execution is affected by tracking error and safety limits.

\paragraph{Construct validity}
The common \SI{1}{\second} window improves cross-model comparability but can omit later braking or avoidance. Feasibility thresholds filter obvious anomalies rather than certify vehicle dynamics. The all-correct CIA exposes binding failure but can undervalue partially correct understanding. IVCS weights are specific to the present evaluation objective and are not universal safety constants.

\paragraph{External validity}
The present paired set focuses on Cut-in and VRU crossing across three domain implementations. It does not yet cover intersection negotiation, merging, occluded crossing, adverse weather, high-speed events, or continuous multi-agent interaction. Closed-track video includes real imaging and execution disturbance but not the natural traffic distribution of public roads. The macro ordering is therefore specific to the evaluated models, scenarios, and domain implementations.

\paragraph{Model and interface validity}
OpenEMMA, LLaViDA, and Alpamayo-R1 differ in training data, foundation model, video history, coordinates, and output interface. Cross-paradigm results characterize the whole system and do not identify the causal contribution of an internal module. Foundation-model services may also change through server-side versions and routing. Reproduction requires frozen model identifiers, prompts, preprocessing, temperatures, and parsers together with raw outputs.

\paragraph{Evaluator validity}
The combined rule and constrained-LLM extractor may miss synonyms, bind negation incorrectly, or confuse the described subject. An anonymized, double-reviewed subset should quantify slot-extraction agreement, false positives, and false negatives. Any API credential used in evaluation must be removed and rotated before release.

\subsection{Future work}
\label{subsec:future-work}

Four extensions are priorities. First, the scenario graph should add unprotected turns, merging, abrupt lead-vehicle braking, occluded crossing, roundabouts, and continuous multi-vehicle interaction. Second, transfer ablations and qualification evidence should compare direct compilation without structure, object-relation-only structure, and full SSP. Third, broader parameter coverage and closed-loop execution should connect video-conditioned short-horizon trajectories to realized vehicle response, minimum separation, and risk duration in CARLA and on the proving ground. Fourth, a public SSP data and tool release should include source videos, structured scene cards, CARLA configurations, proving-ground task templates, raw VLA outputs, labels, and evaluation code so that domain-conditioned findings can be tested across models and sites.

\section{Conclusion}
\label{sec:conclusion}

This study addresses a central confound in autonomous-driving VLA evaluation: changes in data domain are often accompanied by uncontrolled changes in scenario content. SSP resolves this problem by defining the interaction event independently of its domain realization. Starting from a synthetic long-tail source, a VLM-assisted and human-verified event specification records the topology, participant roles, dominant relative motion, conflict relationship, passing order, admissible response, and event phases that must remain preserved. CARLA and proving-ground assets are then constructed under platform-specific constraints and admitted to evaluation only after explicit transfer qualification. The resulting Synthetic, Simulation, and Physical videos therefore form an auditable event-matched evidence set rather than three independently selected examples.

SSP also provides a common behavior evaluation for VLA language and action. Evidence-grounded slots represent object, position, motion, conflict, and action, while heterogeneous trajectories are transformed to a common ego-centric frame and a \SI{1}{\second} interval. Output validity, effective semantic accuracy, critical interaction, explicit action, trajectory quality, text--trajectory consistency, and sustained risk response distinguish well-formed but incorrect output, correct language not implemented by a trajectory, and mutually consistent but jointly incorrect behavior.

The evaluation produces three findings. First, Physical, Simulation, and Synthetic obtain macro IVCSs of 0.325, 0.291, and 0.259, respectively, but Simulation leads Cut-in at 0.340 and Physical leads VRU crossing at 0.374; domain effect is scenario conditioned rather than monotone. Second, Alpamayo-R1, OpenEMMA, and LLaViDA attain IVCSs of 0.405, 0.338, and 0.131, with the Alpamayo-R1 advantage spanning effective semantics, critical-interaction recognition, and trajectory quality. Third, Qwen3-VL-30B-A3B attains a TQSR of 0.630 and IVCS of 0.398 with approximately 3B active parameters, exceeding 32B Dense at 0.426/0.366, while the larger 235B-A22B is not monotonically superior. MoE benefit should therefore be interpreted as active-parameter efficiency and a capability trade-off, not a simple scaling law.

From an engineering perspective, SSP supports high-coverage screening in Synthetic data, controlled reconstruction and diagnosis in Simulation, and targeted confirmation in Physical tests. It does not claim that nonphysical data replace physical validation or that Physical data are an absolute ground truth for every metric. The present conclusions remain bounded by the paired scenario set, three domain implementations, and heterogeneous model interfaces. Broader scenario coverage, closed-loop response analysis, and public transfer audits will extend this evidence base.

\section*{CRediT authorship contribution statement}
\textbf{Haojie Feng}: Conceptualization, Methodology, Software, Formal analysis, Investigation, Data curation, Visualization, Writing -- original draft. 
\textbf{Peizhi Zhang}: Conceptualization, Methodology, Supervision, Project administration, Writing -- review \& editing. 
\textbf{Xinrui Zhang}: Methodology, Software, Validation, Formal analysis, Data curation, Visualization, Writing -- review \& editing. 
\textbf{Zhuoren Li}: Investigation, Validation, Data curation, Resources. 
\textbf{Junpeng Huang}: Methodology, Software, Formal analysis, Visualization, Writing -- review \& editing. 
\textbf{Xiurong Wang}: Investigation, Validation, Resources. 
\textbf{Dongxiao Yin}: Investigation, Validation, Resources. 
\textbf{Yuxiang Zhang}: Investigation, Validation, Resources. 
\textbf{Junfan Zhu}: Methodology, Formal analysis, Validation, Writing -- review \& editing. 
\textbf{Lu Xiong}: Conceptualization, Supervision, Project administration, Funding acquisition, Writing -- review \& editing.

\section*{Declaration of competing interest}
The authors declare that they have no known competing financial interests or personal relationships that could have appeared to influence the work reported in this paper.

\section*{Acknowledgments}
This study was supported by the National Natural Science Foundation of China (Grant No. 52325212).

\section*{Data availability}
Data will be made available on request.

\section*{Declaration of generative AI and AI-assisted technologies in the manuscript preparation process}
During the preparation of this work, the authors used ChatGPT to assist with English translation and language editing. The authors reviewed and edited the output as needed and take full responsibility for the content of the published article. 

\renewcommand{\bibfont}{\fontsize{7.5}{8.5}\selectfont}
\renewcommand{\UrlFont}{\ttfamily}
\bibliographystyle{elsarticle-num}
\bibliography{references}

\end{document}